%% file: iclr2027_conference.tex
\documentclass{article} 
\usepackage{iclr2027_conference,times}

\input{math_commands.tex}

\usepackage{hyperref}
\usepackage{url}

\usepackage{booktabs}
\usepackage{multirow}
\usepackage{graphicx}
\usepackage{amssymb}
\usepackage[table]{xcolor}
\usepackage[normalem]{ulem}
\useunder{\uline}{\ul}{\underline}
\usepackage{rotating}
\usepackage{caption}
\newcommand{\fitwidth}{\ifdim\width>\linewidth\linewidth\else\width\fi}

\usepackage{cleveref}
\usepackage{wrapfig}
\usepackage{amsmath,amssymb}
\usepackage{algorithm}
\usepackage{algpseudocode}
\usepackage[table]{xcolor}
\usepackage{seqsplit}
\usepackage{tcolorbox}
\tcbuselibrary{breakable}
\usepackage[normalem]{ulem}
\usepackage{graphicx}
\usepackage{multirow}  
\usepackage{fvextra}
\usepackage{adjustbox}
\usepackage{subcaption}

\usepackage{pifont}

\title{RetroPlanner: Multi-Route Expansion and Multi-Objective Reasoning for LLM-Based Retrosynthesis Planning}

\author{
Ye Rin Kim\textsuperscript{1} \hspace{0.7em}
Yongjun Jeong\textsuperscript{1} \hspace{0.7em}
Seongjun Yun\textsuperscript{1} \hspace{0.7em}
Hanbum Ko\textsuperscript{1} \hspace{0.7em}
Soohyun Oh\textsuperscript{1} \\[1mm]
\textbf{
Sungbin Lim\textsuperscript{2}\textsuperscript{*} \hspace{1.2em}
Sungwoong Kim\textsuperscript{1}\textsuperscript{*}
} \\[1mm]
\textsuperscript{1}Korea University, Department of Artificial Intelligence \\
\textsuperscript{2}Korea University, Department of Statistics \\
{\small\texttt{\{yerin-kim,\,dydrkfl0608,\,seongjunyun,\,hanbumko95,\,soohyun\_oh\}@korea.ac.kr}} \\
{\small\texttt{sungbin@korea.ac.kr}\hspace{1.2em}\texttt{swkim01@korea.ac.kr}}
}

\iclrfinalcopy 
\begin{document}

\maketitle

\renewcommand{\thefootnote}{\fnsymbol{footnote}}
\footnotetext{%
  \textsuperscript{*}Corresponding authors. \
  Code is available at \url{https://github.com/KU-AGI/RetroPlanner}.%
}
\renewcommand{\thefootnote}{\arabic{footnote}}

\input{contents/00_Abstract}
\input{contents/01_Introduction}
\input{contents/02_Method}
\input{contents/03_Experiments}

\input{contents/04_Ablation_Analysis}
\input{contents/05_Limitations_Conclusion}

\newpage

\subsection*{AI use statement}

In this work, we used generative AI tools for writing assistance, literature retrieval and discovery, research ideation and implementation support, and the generation of synthetic training data.
For writing, we used LLMs to refine phrasing and improve the clarity of our drafts, and we reviewed and edited all resulting text.
For retrieval and discovery, we used LLM-based search to locate candidate related work, and we read and verified every cited paper.
For research ideation and execution, we used LLMs to discuss design alternatives for the planning agent and to assist in writing experiment and analysis code, which we reviewed and tested before use.
For synthetic data generation, LLMs were used to construct the supervised fine-tuning corpus of reasoning trajectories, as described in~\Cref{sec:method_learn}; every trajectory was verified against the recorded search graph, and only traces passing this verification were used for training.
We have reviewed all AI-assisted work and take responsibility for the final content of this work, including text and artifacts produced with the aid of generative AI.

\subsection*{Ethics statement}

This work does not involve human subjects, personally identifiable information, or the release of new experimental data.
All experiments use publicly available reaction datasets and benchmarks, and all models and building-block catalogs are used in accordance with their licenses.
Retrosynthesis planning is a domain where a method that proposes practical routes could in principle be applied to harmful target molecules.
Our method does not introduce a new reaction generator or expand the underlying candidate space; instead, it changes how routes proposed by existing single-step models are compared and reordered during search.
The proposed routes are computational suggestions evaluated using model-based signals and require expert review and experimental validation before any laboratory use.

\subsection*{Reproducibility statement}

We provide the search formulation, reasoning schema, and training procedure of \textsc{RetroPlanner} in \Cref{main:method}, and the evaluation protocol and ablation settings in \Cref{main:experiments}.
The appendix describes the training data construction, input prompt and output format, hyperparameters, and evaluation details in \Cref{appx:training_data_construction,appx:reasoning_prompt,appx:evaluation_details}.
All datasets, the single-step model, the base model, and the scoring tools we use are publicly available.
Additional details and code are provided in the supplementary material.


\subsubsection*{Acknowledgments}
This work was also supported by the National Research Foundation of Korea (NRF) grant funded by the Korean government (MSIT) (No. RS-2026-25613293, Project Title: Physics- and Mathematics-Informed Causal AI Models for Robust Analysis and Prediction under Geometry, Boundary-Condition, Parameter, and Multi-Environment Shifts, 20\%), and was also supported by the Institute of Information \& Communications Technology Planning \& Evaluation(IITP) grant funded by the Korea government(MSIT) (No. RS-2019-II190079, Artificial Intelligence Graduate School Program(Korea University), 20\%), (IITP-2026-RS-2025-02304828, artificial intelligence star fellowship support program to nurture the best talents, 20\%), and (No. RS-2026-25507543, Development of AI Co-Scientist based on Scientific Causal World Model, 20\%), and by the Advanced GPU Utilization Support Program funded by the Government of the Republic of Korea (Ministry of Science and ICT).

\newpage

\bibliography{iclr2027_conference}
\bibliographystyle{iclr2027_conference}

\newpage

\input{contents/99_Appendix}

\end{document}

%% file: math_commands.tex
\usepackage{amsmath,amsfonts,bm}

\def\Figref#1{Figure~\ref{#1}}

\def\eqref#1{equation~\ref{#1}}

\def\1{\bm{1}}

\DeclareMathAlphabet{\mathsfit}{\encodingdefault}{\sfdefault}{m}{sl}
\SetMathAlphabet{\mathsfit}{bold}{\encodingdefault}{\sfdefault}{bx}{n}



%% file: contents/00_Abstract.tex
\begin{abstract}
Multi-step retrosynthesis planning is often considered solved once a route to purchasable starting materials is found, but such a route should also be practical in real-world chemistry.
Recent LLM-based planners formulate this process as sequential decision-making over an AND-OR graph, yet typically expand a single molecule per turn and are optimized for solvability alone, leaving alternative routes uncompared across the multiple objectives that determine practicality before the search budget is committed.
We introduce \textsc{RetroPlanner}, an LLM-based planner that jointly compares multiple routes to steer the search toward higher synthesizability, as measured by reaction plausibility, round-trip consistency, and starting-material price.
\textsc{RetroPlanner} is trained to reason over chemical evidence: structural progress from the target molecule, bond changes and reaction classes, where candidate routes diverge from one another, and how they rank across multiple objectives.
Experiments on multi-step retrosynthesis benchmarks demonstrate that \textsc{RetroPlanner} recovers more alternative routes than prior LLM-based planners while improving synthetic route quality and maintaining a high success rate.
\end{abstract}

%% file: contents/01_Introduction.tex
\section{Introduction}
\label{sec:introduction}

\input{figtabs/01_motivation_examine}

Multi-step retrosynthesis planning is a fundamental task in organic synthesis and drug discovery that identifies a sequence of reactions connecting a target molecule to starting materials available in a set of purchasable building blocks~\citep{corey1969computer}.
This task is commonly formulated as a search over an AND-OR graph~\citep{schreck2019learning,chen2020retro}, in which a single-step retrosynthesis model proposes candidate precursors and a planner determines the next molecule to expand.
A planner's effectiveness depends on how it explores the search space within the available expansion budget and on how it maintains the quality of the routes it recovers.
Conventional planners control the search procedure using predefined algorithms~\citep{segler2018planning,kishimoto2019depth,xie2022retrograph} or learned value functions that estimate the future cost from a candidate~\citep{liu2023retrosynthetic,han2022gnn}, allocating the budget by a static rule aimed only at reaching building blocks.
Recent studies turned to treating route practicality as an explicit objective, formulating planning as a multi-objective search over criteria beyond success rate~\citep{hastedt2026feasible}, with benchmarks following the same shift by asking whether each reaction along a route has synthetic precedent~\citep{zagribelnyy2026ursa}.

Large language models have recently been used as agentic planners that interact with the single-step model through a sequence of tool calls~\citep{liu2026retro,zhu2026retroagent,changretrorchestrator}.
Existing agents may attach multiple reaction candidates when expanding a molecule, but each turn still commits the search to a single frontier molecule.
This makes it difficult to compare competing routes under the limited search budget, even though the candidates at the other frontier molecules could be considered in the same turn.
Such a comparison starts at the step level, where each decision determines the synthesizability of the resulting route: the chosen reaction should be plausible under the distribution of known reactions~\citep{Lowe2017}, as measured by the reaction filter models used in multi-step search software~\citep{genheden2020aizynthfinder,tu2025askcos,maziarz2025re}.
In addition, each decision should consider whether the predicted precursors reproduce the product under a forward model, known as round-trip consistency~\citep{schwaller2020predicting,lin2022improving}, and at the route level, whether the route reaches inexpensive starting materials, as estimated by a price predictor~\citep{hastedt2025molprice}.

To motivate our approach, we examine the routes available within a retrosynthetic search space built for targets from the subset of the USPTO-190~\citep{chen2020retro} benchmark that are solvable within a depth of five, where R-SMILES~\citep{zhong2022root} returns its top-4 reaction candidates for each molecule.
We compare these routes from four different views: (1) \emph{Agent Expanded}: routes successfully found by RetroAgent~\citep{zhu2026retroagent}, (2) \emph{Exposed, but Unexpanded}: routes whose molecules are opened by the agent during search but are not further expanded, (3) \emph{Unexposed}: routes recovered only by exhaustively expanding that space with the same candidates, and (4) \textsc{RetroPlanner}: routes recovered by our planner through joint multi-route expansion and multi-objective comparison.
As shown in~\Cref{fig:motivation}, many routes that dominate the agent's routes in reaction plausibility, round-trip consistency, and starting-material price are already within its reach, among those the agent itself opened and left unexpanded.
Among the 42 targets solved by the agent in this 60-target subset, the agent's route is dominated by some route in the search space on 78.6\% of them, and on 64.3\% that dominating route is already among those exposed but left unexpanded.
Recovering these routes benefits from two components: opening several molecules within a turn places their candidates in view at once, and comparing candidates across the three objectives helps retain synthesizable routes.
\textsc{RetroPlanner} is designed to realize this behavior by jointly expanding multiple frontier molecules and comparing their candidate routes in one multi-objective decision.

In this paper, our contributions are summarized as follows.

\begin{itemize}
    \item \textbf{Multi-Route Expansion.}
    We formulate frontier expansion as a joint decision over multiple molecules.
    \textsc{RetroPlanner} opens several frontier molecules within the same turn, so that competing routes are evaluated against one another, yielding more diverse routes. 
    \item \textbf{Multi-Objective Reasoning.}
    We define route quality as a reasoning objective over reaction plausibility, round-trip consistency, and starting-material price, so that each expansion decision explicitly states the structural change of the molecule, examines how generally applicable the reaction is at the bond and reaction-class level, and specifies where the candidate stands among the routes already explored and which objective separates it from the alternatives.
    \item \textbf{Diverse and Synthesizable Route Discovery.}
    \textsc{RetroPlanner} recovers substantially more alternative routes than prior LLM-based planners while improving their synthesizability across decision signals, approaching dedicated multi-objective search methods while maintaining a high success rate.
\end{itemize}

%% file: figtabs/01_motivation_examine.tex
\begin{wrapfigure}{r}{0.48\textwidth}
    \vspace{-20pt}
    \centering
    \includegraphics[width=0.40\textwidth]{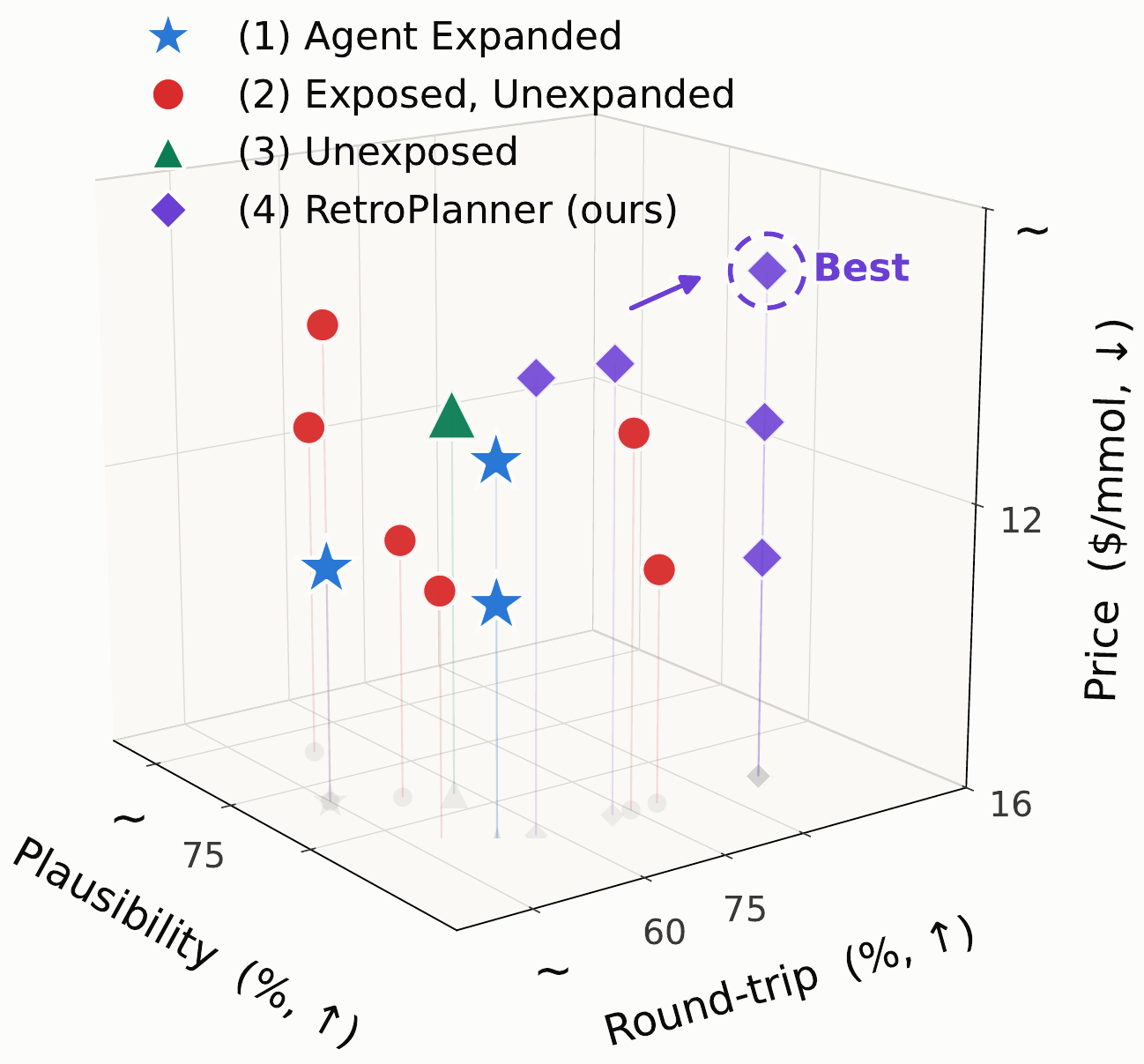}
    \caption{
    Since routes that dominate the agent's returned routes on reaction plausibility, round-trip consistency, and starting-material price are often already exposed during search but left unexpanded, \textsc{RetroPlanner} recovers them by jointly expanding multiple frontier molecules and comparing them in a single multi-objective decision per turn.
    }
    \label{fig:motivation}
    \vspace{-10pt}
\end{wrapfigure}

%% file: contents/02_Method.tex
\section{Method}
\label{main:method}

\subsection{Problem Formulation}
We first formulate multi-step retrosynthesis planning as a sequential search over an AND--OR graph that evolves over search turns. 
Let $\mathcal{M}$ be the set of all molecules and $\mathcal{R}$ the set of all reactions, and let $t=0,1,\ldots$ index the search turn, with $t=0$ the initial state containing only the target molecule $m_0$.
At turn $t$, the current search graph is denoted by:
\begin{equation}
    \mathcal{G}_t = (\mathcal{V}_{t}^{M}, \mathcal{V}_{t}^{R}, \mathcal{E}_t),
\end{equation}
where $\mathcal{V}_{t}^{M}\subset \mathcal{M}$ and $\mathcal{V}_{t}^{R}\subset \mathcal{R}$ denote the sets of molecule nodes $m\in\mathcal{M}$ (OR nodes) and reaction nodes $r \in \mathcal{R}$ (AND nodes) discovered up to turn $t$, respectively. 
$\mathcal{E}_t$ is the set of directed edges between molecules and reactions at turn $t$. 
We keep $\mathcal{G}_t$ a directed acyclic graph (DAG), where a molecule appears as a single node shared by every route that contains it.
\textsc{RetroPlanner} therefore observes the explored routes on the same graph, as shown in \hyperref[fig:main]{Figure~\ref*{fig:main}(c)}, so a candidate can be compared against the routes already explored.

Let $\mathcal{BB}\subset \mathcal{M}$ denote the set of purchasable building-block molecules, and $\operatorname{ch}(\cdot)$ denote the children of a node. Standard AND--OR semantics then give
\begin{align}
    \texttt{SOLVED}_t(m) &\iff m\in\mathcal{BB} \;\vee\;
        \exists\, r\in\operatorname{ch}_t(m):\ \texttt{SOLVED}_t(r), \\
    \texttt{SOLVED}_t(r) &\iff
        \forall\, m'\in\operatorname{ch}_t(r):\ \texttt{SOLVED}_t(m'),
\end{align}
so a molecule is solved once any one of its reactions is, whereas a reaction needs its precursors $m'$ to be solved.
We define the set of frontier molecules at turn $t$ as
\begin{equation}
    \mathcal{O}_t = \{\, m\in\mathcal{V}_{t}^{M} : \neg\texttt{SOLVED}(m) \,\}.
\end{equation}
$\mathcal{O}_t$ is the set of molecules that are not yet solved and therefore remain candidates for expansion.
Since opened molecules retain their candidates, the agent can revisit them at later turns and expand several at once, yielding alternative routes through the same subgraph.

\subsection{Overview of \textsc{RetroPlanner}}

\input{figtabs/03_method_graph}

\paragraph{LLM planner with environment.}

At turn $t$ the planner observes the search graph $\mathcal{G}_t$ and acts according to LLM policy $\pi_{\theta}(\cdot \mid \mathbf{h}_t)$ with history $\mathbf{h}_t=(\mathcal{G}_0, \tilde{a}_0, \cdots, \mathcal{G}_{t-1}, \tilde{a}_{t-1}, \mathcal{G}_t)$. 
An action $\tilde{a}_t$ consists of decisions, where $z$ denotes a reasoning trace preceding the corresponding decision,
\begin{equation}
    \tilde{a}_t = (z_t^{\texttt{open}}, a_t^{\texttt{open}}, z_t^{\texttt{rank}}, a_t^{\texttt{rank}}, z_t^{\texttt{done}}, a_t^{\texttt{done}}).
\end{equation}
The three decisions correspond to the \texttt{open}, \texttt{rank}, and \texttt{done} stages of a turn. 
The episode terminates when the planner takes the terminal action $a^{\texttt{final}}$.
They are generated autoregressively so that every decision is conditioned on history produced earlier in the turn, including the trace that precedes it.

\paragraph{Multi-Route Expansion.}
Existing agentic planners~\citep{liu2026retro,zhu2026retroagent} typically operate on a single frontier molecule per turn: the top-$K$ reaction candidates branch only within that molecule, never across molecules, so the graph widens only locally as in \hyperref[fig:main]{Figure~\ref*{fig:main}(a) and (b)}.
We instead allow \emph{set-valued} decisions over multiple molecules $m\in \mathcal{O}_t$ to simultaneously explore diverse routes and balance multi-objective trade-offs within a single turn.
In the \texttt{open} stage, the agent jointly selects a subset of molecules to expand,
\begin{equation}
    a_t^{\texttt{open}} \subseteq \mathcal{O}_t,
\end{equation}

and invokes \texttt{open}; for each selected $m$, the environment provides $f_K(m)$, the top-$K$ reaction candidates proposed by the single-step model $f(\cdot)$.
Prior planners are restricted to $|a_t^{\texttt{open}}|=1$ and thus advance a single route per turn.

In the \texttt{rank} stage, the action $a^{\texttt{rank}}_t$ orders reaction candidates of each opened molecule under the multi-objective criterion and determines how many of the top-ranked reactions to expand for each opened molecule.
\begin{equation}
\label{eq:method_rank}
    a^{\texttt{rank}}_t \in \prod_{m \in a^{\texttt{open}}_t}
    \mathcal{S}_{\le L}\bigl(f_K(m)\bigr),
\end{equation}
where $\mathcal{S}_{\le L}\bigl(f_K(m)\bigr)$ denotes the ordered tuples of one to $L$ distinct candidates from $f_K(m)$: $K$ limits how many candidates are considered per molecule, and the expansion limit $L$ caps how many of them are expanded.
In our setting we use $K=10$ and $L=3$.

The environment attaches every selected reaction with its precursors to the graph, so several routes grow within a single turn.
In the \texttt{done} stage, $a^{\texttt{done}}_t$ commits the routes the agent judges to be complete; the molecules on a committed route are solved and therefore leave $\mathcal{O}_t$, and the search continues on the routes that remain.

\paragraph{Multi-Objective Reasoning.}
Expanding multiple routes exposes alternatives with different quality trade-offs.
We consider synthesizability through three objectives: reaction plausibility, round-trip consistency, and starting-material price.
We therefore train the agent to reason over these objectives rather than optimizing only for success rate. 
Rather than imposing a fixed weighting across the search, different objectives can guide individual expansion decisions, collectively shaping the trade-off among the three objectives at the route level.

Every reaction candidate $r \in f_K(m)$ is therefore displayed with its plausibility $\mathrm{p}$, round-trip consistency $\mathrm{rt}$, and starting-material price $\mathrm{c}$, so that candidates are comparable both within a molecule and across the opened molecules.
For any route $\rho$ in the current graph, route quality is the multi-objective vector
\begin{equation}
    \mathbf{u}(\rho) = \bigl[\, u_{\mathrm{p}}(\rho),\, u_{\mathrm{rt}}(\rho),\, u_{\mathrm{c}}(\rho) \,\bigr],
\end{equation}
which the agent uses to compare routes against one another during the search, whether they are solved or still being explored.
This reasoning takes place in the rank trace $z_t^{\texttt{rank}}$, where the agent compares candidates along the three objectives before committing to the ordered selection $a_t^{\texttt{rank}}$.
The agent also considers auxiliary signals, such as single-step model confidence $\mathrm{q}$ and structural descriptors of molecules in the current graph, which inform the comparison of candidates and partial routes.

\input{figtabs/03_method_reasoning} 

\subsection{Reasoning Schema}
\label{sec:method_learn}

\paragraph{Evidence-Grounded Reasoning.}
We require the agent to specify the evidence behind each choice in reasoning field.
First, the \texttt{MOLECULE} field compares the molecule at the current depth with the target, characterizing how far its structure has been simplified through changes in scaffold, size, rings, stereochemistry, and precursor-fragment decomposition.
Second, the \texttt{DISCONNECTIONS} field identifies the bond changes and associates the transformation with a reaction class when available, allowing each candidate to be recognized as a known transformation.
Third, the \texttt{DECIDING} field references numerical signals provided on the search graph, such as reaction plausibility, round-trip consistency, and starting-material price, so that candidates and the routes they extend can be compared against all explored partial routes, with every selected candidate carrying two subfields, \texttt{diverging from} and \texttt{ordered by}.
Detailed examples for all three fields are provided in~\Cref{appx:reasoning_prompt}.

\paragraph{Comparative Grounding across Routes and Objectives.}
In the \texttt{diverging from} subfield, the agent refers to the explored route or the competing candidate in the same turn that is most relevant to the candidate, based on the acquired chemical evidence and the search history.
It also notes when the molecule has not been expanded by any explored route.
In the \texttt{ordered by} subfield, the agent identifies the objective most decisive for ordering the candidate among its competing alternatives in the current decision. 
Different objectives can therefore be decisive at different steps along the same route.
The primary objectives are reaction plausibility $\mathrm{p}$, forward round-trip consistency $\mathrm{rt}$, and starting-material price $\mathrm{c}$, the last on a normalized log scale.
Single-step confidence $\mathrm{q}$ serves as an auxiliary signal when the leading candidates are nearly tied on the primary objectives.

\subsection{Training}

\paragraph{Dataset.}

We construct the training set from the PaRoutes~\citep{genheden2022paroutes} training split.
Each target is searched once with Retro*-0~\citep{chen2020retro}, using R-SMILES~\citep{zhong2022root} as a template-free single-step model to sample diverse retrosynthetic candidates while reducing dependence on template coverage.
We Pareto-rank the searched routes over the three synthesis objectives and merge a diverse subset of high-ranking routes at their shared intermediates into a single directed acyclic graph, with the selection procedure detailed in~\Cref{appx:training_data_construction}.
We then replay this graph breadth-first in a deterministic simulation of the planning environment, which converts each step into one of the planner's actions: \texttt{open}, \texttt{rank} with the number of committed candidates, or \texttt{done}, and ends the episode with \texttt{final}.
We use Qwen3.8-27B~\citep{qwen38} as a teacher model to generate reasoning traces for these actions, verified as described in~\Cref{sec:method_learn}.

\paragraph{Training with Verified Reasoning.}
During training corpus construction, the structured reasoning schema enables direct verification against the search graph, as illustrated in \hyperref[fig:trace]{Figure~\ref*{fig:trace}(b)}.
Every numerical value cited in the reasoning trace must match what the graph records.
For example, a reasoning trace reporting $\texttt{value}: 0.82$ for a selected candidate whose recorded plausibility is $0.67$ fails verification.
For each selected candidate, the route cited in \texttt{diverging from} must exist in the graph and differ from the selected candidate as stated, and the objective cited in \texttt{ordered by} must separate the candidate from its competitors in the stated objective.
A failed trace is returned to the teacher LLM with the detected inconsistencies as repair feedback, or regenerated when the defects cannot be localized.
When no trace passes verification, the action has no reasoning trace, so only verified reasoning is used for supervised fine-tuning.
Given a search-state input $x$ and its verified reasoning-and-action sequence $y$, we optimize the standard autoregressive objective: 
$\mathcal{L}_{\mathrm{SFT}}(\theta)=-\sum_{t=1}^{|y|}\log p_{\theta}(y_t \mid x, y_{<t})$.

%% file: figtabs/03_method_graph.tex
\begin{figure}[t]
    \centering
    \scalebox{1}[1]{\includegraphics[width=\linewidth]{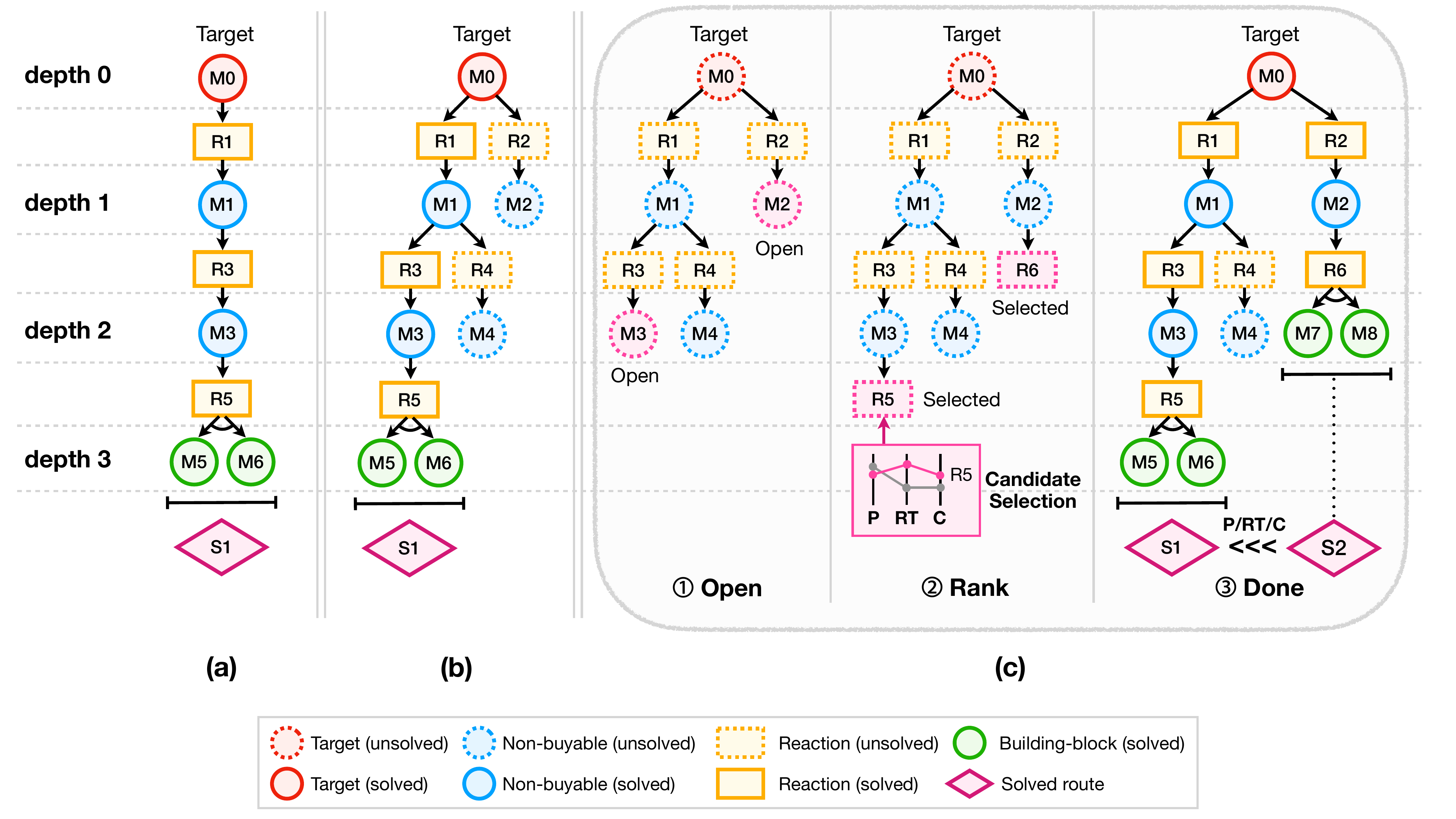}}
    \caption{
    Planning on the same target under three expansion patterns, corresponding to the strategies analyzed in~\Cref{abl:expansion_strategy}.
    (a) Single-route expansion extends one route per turn, as in Retro-R1.
    (b) Sequential multi-route expansion grows several routes, but each turn opens and ranks one frontier molecule, as in RetroAgent.
    (c) Parallel multi-route expansion, used by \textsc{RetroPlanner}, runs three stages per turn: \texttt{open} selects multiple molecules from the frontier set $\mathcal{O}_t$ to expand at once, \texttt{rank} jointly orders the reaction candidates of the opened molecules for expansion through multi-objective reasoning over reaction plausibility (P), round-trip consistency (RT), and starting-material price (C).
    \texttt{done} commits routes whose leaves are all in the building-block set $\mathcal{BB}$.
    }
    \label{fig:main}
\end{figure}

%% file: figtabs/03_method_reasoning.tex
\begin{figure}[t]
    \centering
    \scalebox{1}[1]{\includegraphics[trim={0 10pt 0 10pt}, clip, width=\linewidth]{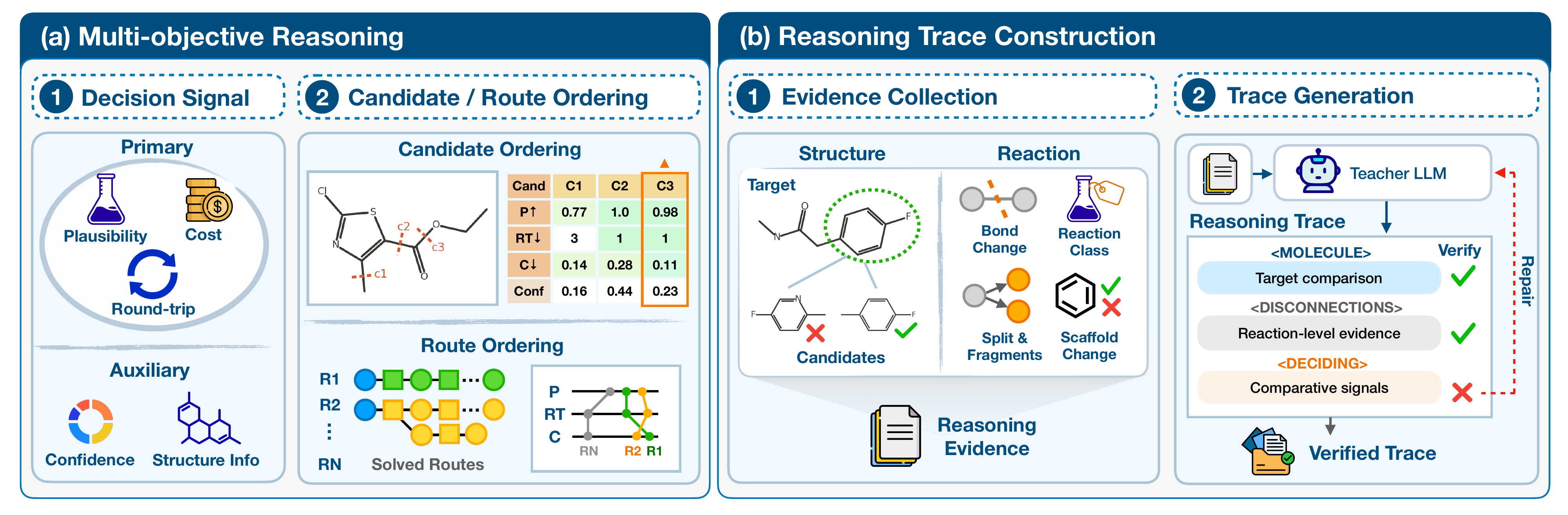}}
    \caption{
    \textbf{(a)} Multi-objective reasoning. Reaction candidates are compared on three primary decision signals and auxiliary signals, which are referenced to order the candidates of each molecule. \textbf{(b)} Reasoning trace construction. Structural and reaction-level evidence is collected for the target and its candidates; a teacher LLM writes the \texttt{MOLECULE} / \texttt{DISCONNECTIONS} / \texttt{DECIDING} trace from that evidence, and each field is verified against the search graph, with detected inconsistencies returned as repair feedback until the trace passes.
    }
    \label{fig:trace}
\end{figure}

%% file: contents/03_Experiments.tex
\section{Experiments}
\label{main:experiments}

\subsection{Experimental Setup}

\paragraph{Baselines.}

We compare our method against two groups of baselines.
First, we include conventional multi-step search algorithms, MCTS~\citep{segler2018planning}, Retro*-0, Retro*~\citep{chen2020retro}, and MORetro*~\citep{hastedt2026feasible}, evaluated under the same single-step retrosynthesis model and search budget $N_B$.
All main results use $N_B=50$, while the ablation study raises this to $N_B=300$ so that each variant is compared under a budget large enough not to be the limiting factor.
For a controlled comparison, we replace the confidence score that drives node selection in these algorithms with an equal-weight scalarization of reaction plausibility, round-trip consistency, and starting-material price.
For MORetro*, we retain its original Bayesian-optimization (BO) sampling procedure over objective weights.
Second, we include trained LLM-based planners such as Retro-R1~\citep{liu2026retro} and RetroAgent~\citep{zhu2026retroagent}.
\textsc{Retroplanner} is initialized from GPT-OSS-20B~\citep{openai2025gptoss120bgptoss20bmodel} and supervised fine-tuned on our training corpus to learn multi-route expansion and multi-objective reasoning.

\paragraph{Evaluation Setup.}
We evaluate on the in-distribution USPTO-190~\citep{chen2020retro} and the out-of-distribution ChEMBL-1000~\citep{liu2023retrosynthetic}.
Search baselines are run once per budget cell under the common expansion budget.
LLM-based planners instead run multiple rollouts that consume a cumulative expansion budget until exhausted.
Following reported settings~\citep{liu2026retro,zhu2026retroagent}, Retro-R1 searches within a single continuous rollout over multiple rounds, whereas RetroAgent restarts failed rollouts until a route is found.
\textsc{RetroPlanner} ends each rollout at its first complete route and restarts, spending the remaining budget on alternative routes.
Unlike the baselines, it does not shuffle the environment and instead samples its reasoning and decisions at temperature 0.6, so that variation across rollouts comes from the agent itself.
Since these settings differ across baselines, we further train each expansion strategy on the same backbone and search mechanism, as detailed in~\Cref{abl:expansion_strategy}.

\paragraph{Metrics.}

We group metrics into \textbf{Search Outcome} and \textbf{Route Quality}.
\emph{Success Rate (Succ.)} is the percentage of targets for which a route reaching purchasable building blocks from eMolecules\footnote{\url{https://downloads.emolecules.com/}} is found within the expansion budget, and \emph{Routes} is the number of routes returned per solved target.
For route quality, we report the composite-best route recovered by each method for each target, selected jointly over reaction \emph{Plausibility (Plaus.)}, \emph{Round-Trip (RT)} consistency, and starting-material \emph{Price}.
\emph{Better-Than-Ref (BTR)} is the fraction of targets for which the selected route attains a higher composite score than the best published reference route over the same three objectives, with unsolved targets counted as failures.
Additionally, \emph{1stCuts} counts the distinct first-cut precursor sets recovered per target and \emph{1stCutDiff} is the fraction of route pairs with different first-cut precursor sets, both measuring how broadly a planner explores alternative branches at the initial retrosynthetic decision.
Further evaluation details are provided in~\Cref{appx:evaluation_details}.

\subsection{Main Results}

\noindent\textbf{Q1. Does the planner efficiently recover diverse, high-quality routes under a constrained search budget?}

We evaluate all methods under an expansion budget of $N_B=50$, focusing on how effectively each planner allocates a limited search budget toward solvable and high-quality routes under the three evaluation objectives.

\input{tables/04_main_results}

\paragraph{Search Outcome.}

As shown in~\Cref{tab:main_results}, LLM-based planners generally achieve higher success rates than conventional search algorithms under the limited search budget, indicating that learned planning policies allocate search effectively toward solvable routes.
On USPTO-190, MORetro* optimizes weighted combinations of the three route-quality objectives, whereas its expansions are governed by sampled objective weights rather than route-level context-aware decision-making, with a success rate 24.22 percentage points lower than that of \textsc{RetroPlanner}.
Among the LLM-based planners, RetroAgent attains the highest success rate on both USPTO-190 and ChEMBL-1000, as its policy is primarily optimized toward reaching a solved route.
However, it does not explicitly account for synthetic route quality during expansion, whereas \textsc{RetroPlanner} incorporates these considerations into its reasoning, resulting in success rates 4.73 and 5.90 percentage points lower, respectively, but substantially more alternative routes with more favorable trade-offs across route-quality objectives.

\paragraph{Route Quality.}

On USPTO-190, \textsc{RetroPlanner} achieves the highest \emph{Better-Than-Ref}, more often recovering solved routes that exceed the reference under the three evaluation objectives, while maintaining strong performance across all three route-quality metrics.
Compared with MORetro*, which explicitly searches over the same three objectives, \textsc{RetroPlanner} achieves a higher success rate while improving reaction plausibility, round-trip consistency, and starting-material price on both benchmarks.
On the other hand, RetroAgent achieves comparable reaction plausibility and starting-material price, but substantially lower round-trip consistency on both benchmarks, whereas \textsc{RetroPlanner} maintains strong performance across all three synthesizability objectives. 
\textsc{RetroPlanner} achieves a more balanced performance across the three objectives and more frequently recovers routes that score above the reference route under the composite metric on USPTO-190, while showing consistent performance across these objectives on ChEMBL-1000, where reference routes are unavailable.

%% file: tables/04_main_results.tex
\begin{table}[!t]
\centering
\caption{
    Main results on multi-step retrosynthesis planning on USPTO-190 and ChEMBL-1000 with search budget $N_B=50$. \emph{Search Outcome} measures whether and how many routes are found, \emph{Route Quality} evaluates the found routes on reaction plausibility, round-trip consistency, and starting-material price. \emph{Better-Than-Ref} is N/A on ChEMBL-1000, which provides no reference routes. Best and second-best results per dataset are shown in \textbf{bold} and \underline{underlined}.
}
\label{tab:main_results}
\resizebox{0.92\textwidth}{!}{%
\begin{tabular}{@{}lcc@{\hspace{8pt}}cccc@{}}
\toprule

\multirow{2}{*}{\textbf{Method}}
& \multicolumn{2}{c}{\textbf{Search Outcome}}
& \multicolumn{4}{c}{\textbf{Route Quality}} \\
\cmidrule(lr){2-3} \cmidrule(lr){4-7}

& \textbf{Success Rate $\uparrow$}
& \textbf{Routes $\uparrow$}
& \textbf{Better-Than-Ref $\uparrow$}
& \textbf{Plausibility $\uparrow$}
& \textbf{Round-Trip $\uparrow$}
& \textbf{Price $\downarrow$} \\
\midrule

\multicolumn{7}{c}{\textbf{USPTO-190}} \\
\midrule

\rowcolor[gray]{0.95}
\multicolumn{7}{c}{\textit{Search Algorithms}} \\
\midrule

MCTS
& 44.74\%
& 16.88
& 27.89\%
& 43.33\%
& 25.88\%
& 391.19 \\

Retro*-0
& 52.63\%
& 29.55
& 34.21\%
& 49.16\%
& 38.99\%
& 328.02 \\

Retro*
& 57.89\%
& \underline{33.34}
& 38.95\%
& 54.93\%
& 41.64\%
& 281.53 \\

MORetro*
& 67.89\%
& \textbf{43.16}
& \underline{52.11\%}
& 66.16\%
& \underline{58.94\%}
& 250.63 \\

\midrule

\rowcolor[gray]{0.95}
\multicolumn{7}{c}{\textit{LLM-based Search}} \\
\midrule

Retro-R1
& 82.11\%
& 2.99
& 38.42\%
& 71.64\%
& 39.34\%
& 197.29 \\

RetroAgent
& \textbf{96.84\%}
& 3.07
& 46.84\%
& \underline{85.85\%}
& 38.16\%
& \textbf{138.83} \\

\textsc{RetroPlanner}
& \underline{92.11\%}
& 7.41
& \textbf{66.84\%}
& \textbf{86.03\%}
& \textbf{61.11\%}
& \underline{167.80} \\

\midrule

\multicolumn{7}{c}{\textbf{ChEMBL-1000}} \\
\midrule

\rowcolor[gray]{0.95}
\multicolumn{7}{c}{\textit{Search Algorithms}} \\
\midrule

MCTS
& 59.30\%
& 28.31
& \text{N/A}
& 56.05\%
& 49.47\%
& 331.59 \\

Retro*-0
& 72.70\%
& \underline{88.83}
& \text{N/A}
& 69.22\%
& 64.40\%
& 249.71 \\

Retro*
& 74.90\%
& \textbf{89.21}
& \text{N/A}
& 71.45\%
& 66.11\%
& 239.78 \\

MORetro*
& 77.80\%
& 26.60
& \text{N/A}
& 75.65\%
& \underline{72.37\%}
& 221.22 \\

\midrule

\rowcolor[gray]{0.95}
\multicolumn{7}{c}{\textit{LLM-based Search}} \\
\midrule

Retro-R1
& 82.90\%
& 2.74
& \text{N/A}
& 74.74\%
& 53.97\%
& 253.53 \\

RetroAgent
& \textbf{95.90\%}
& 5.28
& \text{N/A}
& \textbf{87.86\%}
& 58.35\%
& \underline{207.43} \\

\textsc{RetroPlanner}
& \underline{90.00\%}
& 10.59
& \text{N/A}
& \underline{87.33\%}
& \textbf{74.22\%}
& \textbf{190.89} \\

\bottomrule
\end{tabular}%
}
\end{table}

%% file: contents/04_Ablation_Analysis.tex
\section{Ablation and Analysis}
\label{main:ablation_analysis}

We analyze which components of \textsc{RetroPlanner} contribute to its search behavior and route quality. 
In this section, a rollout continues searching after its first complete route, and rollouts are repeated per target until every variant reaches the same number of single-step calls. 
We separate the expansion strategy into three axes: whether frontier nodes are selected at random or greedily by score alone, whether a single or multiple frontier nodes are expanded per turn, and whether the expansions are performed sequentially or in parallel.
For the reasoning schema, we examine which forms of chemical evidence grounding are useful for planning and how much reasoning over the three synthesis objectives contributes to route diversity and quality.
Since every candidate the planner selects from is proposed by the single-step model, we also replace this model to verify that the planner's gains hold across its variants.

\subsection{Expansion Strategy}
\label{abl:expansion_strategy}

\noindent\textbf{Q2. How does the expansion strategy shape route-set coverage and search efficiency under extended search?}

\input{tables/05_expansion_strategy}

\paragraph{Expansion Strategy Variants.}
For a controlled comparison of expansion strategies on USPTO-190, all methods use the same extended search budget $N_B=300$.
\Cref{tab:expansion_strategy} compares \emph{Multi-Route (Parallel)}, which selects multiple frontier molecules and expands them together within a single turn, against variants that each change one component.
\emph{Random} and \emph{Score-only} replace the learned policy with fixed rules: \emph{Random} selects frontier molecules and reaction candidates uniformly at random, and \emph{Score-only} ranks candidates by an equal-weight scalarization of normalized reaction plausibility, round-trip consistency, and starting-material price.
The other two are separately trained policies that differ from \emph{Multi-Route (Parallel)} only in how frontier molecules are expanded: \emph{Single-Route} expands a single frontier molecule per turn, and \emph{Multi-Route (Sequential)} selects multiple frontier molecules but expands them one per turn instead of together.

\paragraph{Effect of Expansion Strategy.}
\emph{Random} and \emph{Score-only} return fewer than two routes per target.
Scalarizing the three objectives raises \emph{BTR} from 0.280 to 0.672, but scoring each candidate independently does not accumulate alternative routes across the search.
Among the learned policies, selecting multiple frontier molecules broadens the search, as \emph{Multi-Route (Sequential)} raises the number of distinct first cuts from 2.23 to 3.88 over \emph{Single-Route}.
\emph{Multi-Route (Parallel)} recovers fewer routes than \emph{Multi-Route (Sequential)} but keeps a comparable \emph{1stCutDiff}, indicating fewer redundant routes under each first cut, and achieves the highest success rate and \emph{Better-Than-Ref}.
Comparing candidates across frontier molecules within a single turn therefore allocates the search budget toward alternatives that differ from those the graph already holds, rather than toward repeats of the same first cut.

\input{figtabs/05_expantion_time}
Parallel expansion is also cheaper.
As shown in \Cref{fig:expansion_time}, search time grows linearly with the number of molecules opened.
The two multi-route variants open a comparable number of molecules per rollout, 30.4 and 29.6, yet \emph{Multi-Route (Sequential)} spends 13.0\,s per molecule against 10.6\,s for \emph{Multi-Route (Parallel)}, about 23\% more, or 6.5 against 5.1 minutes per rollout.
The gap comes from how much is generated per expansion: per molecule, parallel expansion takes 1.71 turns against 2.30 and 80.2 output tokens against 106.9, so the reasoning over the search graph is shared across the opened molecules rather than repeated for each expansion.
This brings its cost per molecule close to \emph{Single-Route} at 78.5 tokens, while opening twice as many molecules per rollout, 29.6 against 14.8.
Overall, \emph{Multi-Route (Parallel)} obtains its quality gains while searching faster than sequential multi-route search.

\subsection{Reasoning Schema}

\noindent\textbf{Q3. Which components of the reasoning schema contribute to route diversity and quality?}

\input{tables/05_decision_fields}

We construct a controlled hierarchy from the same training trajectories and target actions.
All variants share the same action format and differ only in the reasoning trace that precedes each action.

\paragraph{Effect of Reasoning Schema.}
As shown in~\Cref{tab:reasoning_relation_ablation}, compared with the model trained without explicit reasoning ($L_0$), the full schema ($L_3$) improves all diversity metrics and slightly improves \emph{BTR}, while maintaining a comparable success rate.
Adding molecule-level structural reasoning (\texttt{MOLECULE}, $L_1$) gives the largest single gain, raising \emph{BTR} from 0.720 to 0.769.
Disconnection-level evidence such as bond changes and reaction classes (\texttt{DISCONNECTIONS}, $L_2$) and the retained-candidate record (\texttt{DECIDING}, $L_3$) mainly improve diversity.
\texttt{diverging from} ($L_{3\text{-div}}$) yields the largest number of recovered routes, whereas \texttt{ordered by} ($L_{3\text{-ord}}$) achieves the highest \emph{1stCutDiff}.
Enabling both ($L_3$) maintains high route coverage while yielding the most distinct first cuts and partially recovering the \emph{BTR} lost by \texttt{diverging from} alone.

Molecule-level evidence, which states how far a candidate simplifies the target, improves both the breadth and quality of the retained routes, while disconnection-level evidence, which identifies what a candidate changes, mainly reduces routes that share the same first cut.
With both subfields, the planner retains diverse alternatives that remain grounded in both the search history and the synthesis objectives.
Overall, the reasoning fields shape different aspects of search behavior, and the full schema combines increased route diversity with a high \emph{BTR}.

\subsection{Single-Step and Round-Trip Model Variations}
\input{tables/05_single_step_round_trip_model_variations_2tab}

To assess generalization across reaction models, we vary both the single-step candidate generator and the forward model used for round-trip evaluation on USPTO-190 with $N_B=50$, under the same setting compared in~\Cref{tab:main_results}.
As shown in~\Cref{tab:single_step_comparison}, replacing the template-free R-SMILES used during planner training with the template-based LocalRetro~\citep{chen2021deep}, \textsc{RetroPlanner} retains the highest \emph{Better-Than-Ref} at $53.16\%$ against $43.16\%$ for the strongest baseline, showing that the learned planning policy transfers to an unseen single-step model without additional training.
Similarly,~\Cref{tab:round_trip_variation} shows that replacing ReactionT5v2~\citep{sagawa2025reactiont5} with RXNGraphormer~\citep{xu2025unified} for round-trip evaluation preserves \textsc{RetroPlanner}'s advantage in both \emph{Better-Than-Ref} and \emph{Round-Trip}, which change by only $2.63$ and $2.52$ points, including over MORetro*, which directly optimizes the same route-quality objectives during search.

%% file: tables/05_expansion_strategy.tex
\begin{table}[!t]
\centering
\caption{
    Effect of the expansion strategy on route diversity and quality.
    All variants share the same backbone, GPT-OSS-20B, and differ only in the expansion strategy.
    $^\dagger$Computed only on targets with at least two recovered routes; since Random and Score-only recover few routes, 1.00 reflects complete first-cut disagreement among the available pairs rather than route coverage.
}
\label{tab:expansion_strategy}
\footnotesize
\setlength{\tabcolsep}{2.5pt}
\renewcommand{\arraystretch}{1.08}
\begin{adjustbox}{max width=0.85\linewidth,center}
\begin{tabular}{@{}lccccc@{}}
\toprule
\multirow{2}{*}{\textbf{Method}}
  & \multicolumn{2}{c}{\textbf{Search Outcome}}
  & \multicolumn{2}{c}{\textbf{Diversity}}
  & \textbf{Quality} \\
\cmidrule(lr){2-3}
\cmidrule(lr){4-5}
\cmidrule(lr){6-6}
  & \textbf{Succ. $\uparrow$}
  & \textbf{Routes $\uparrow$}
  & \textbf{1stCutDiff $\uparrow$}
  & \textbf{1stCuts $\uparrow$}
  & \textbf{BTR $\uparrow$} \\
\midrule

Random
& 0.879
& 1.55
& 1.000$^\dagger$
& 1.55
& 0.280 \\

Score-only
& 0.937
& 1.59
& 1.000$^\dagger$
& 1.59
& 0.672 \\

Single-Route
& 0.984
& 6.05
& 0.450
& 2.23
& 0.780 \\

Multi-Route (Seq.)
& \underline{0.995}
& \textbf{9.10}
& \textbf{0.660}
& \textbf{3.88}
& \underline{0.812} \\

Multi-Route (Par.)
& \textbf{1.000}
& \underline{7.75}
& \underline{0.654}
& \underline{3.54}
& \textbf{0.823} \\

\bottomrule
\end{tabular}
\end{adjustbox}
\end{table}

%% file: figtabs/05_expantion_time.tex
\begin{wrapfigure}[20]{l}{0.50\textwidth}
    \centering
    \includegraphics[width=0.47\textwidth]{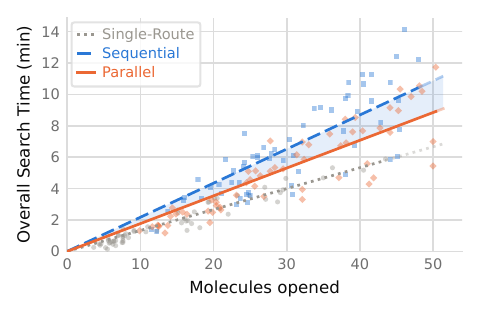}
    \caption{%
    Overall search time versus molecules opened per rollout.
    Although the two multi-route arms open a comparable number of molecules per episode, \emph{Sequential} spends more time per molecule than \emph{Parallel}.
    }
    \label{fig:expansion_time}
\end{wrapfigure}

%% file: tables/05_decision_fields.tex
\newcommand{\dl}[1]{{\scriptsize\,($#1$)}}

\begin{table}[!t]
\centering
\caption{Ablation on the reasoning schema on USPTO-190, with all variants run under the same search budget of at most $N_B=300$.
Schema blocks: \textbf{M}~=~\texttt{MOLECULE},
\textbf{D}~=~\texttt{DISCONNECTIONS},
and \textbf{C}~=~\texttt{DECIDING}.
The \textbf{Rel.} columns indicate the \texttt{diverging from} (\textbf{div}) and
\texttt{ordered by} (\textbf{ord}) subfields of \texttt{DECIDING}.
\textbf{Par.} denotes the parent variant from which each delta is computed.
Best results are in \textbf{bold}.}
\label{tab:reasoning_relation_ablation}

\small
\setlength{\tabcolsep}{3pt}
\renewcommand{\arraystretch}{1.15}

\resizebox{0.88\linewidth}{!}{%
\begin{tabular}{@{}l ccc cc c @{\hspace{8pt}}
r@{}l r@{}l r@{}l r@{}l r@{}l@{}}
\toprule
\multirow{2}{*}{\textbf{Variant}}
& \multicolumn{3}{c}{\textbf{Schema}}
& \multicolumn{2}{c}{\textbf{Rel.}}
& \multirow{2}{*}{\textbf{Par.}}
& \multicolumn{4}{c}{\textbf{Search Outcome}}
& \multicolumn{4}{c}{\textbf{Diversity}}
& \multicolumn{2}{c}{\textbf{Quality}} \\
\cmidrule(lr){2-4}
\cmidrule(lr){5-6}
\cmidrule(lr){8-11}
\cmidrule(lr){12-15}
\cmidrule(lr){16-17}

& \textbf{M} & \textbf{D} & \textbf{C}
& \textbf{div} & \textbf{ord}
&
& \multicolumn{2}{c}{\textbf{Succ. $\uparrow$}}
& \multicolumn{2}{c}{\textbf{Routes $\uparrow$}}
& \multicolumn{2}{c}{\textbf{1stCutDiff $\uparrow$}}
& \multicolumn{2}{c}{\textbf{1stCuts $\uparrow$}}
& \multicolumn{2}{c}{\textbf{BTR $\uparrow$}} \\
\midrule

$L_0$
& -- & -- & -- & -- & -- & --
& 0.989 & \phantom{\dl{\ensuremath{+}0.000}}
& 4.78 & \phantom{\dl{\ensuremath{+}0.00}}
& 0.628 & \phantom{\dl{\ensuremath{+}0.000}}
& 2.79 & \phantom{\dl{\ensuremath{+}0.00}}
& 0.720 & \phantom{\dl{\ensuremath{+}0.000}} \\

$L_1$
& \checkmark & -- & -- & -- & -- & $L_0$
& 0.989 & \dl{\ensuremath{+}0.000}
& 5.49 & \dl{\ensuremath{+}0.71}
& 0.691 & \dl{\ensuremath{+}0.063}
& 3.15 & \dl{\ensuremath{+}0.36}
& \textbf{0.769} & \dl{\ensuremath{+}0.049} \\

$L_2$
& \checkmark & \checkmark & -- & -- & -- & $L_1$
& \textbf{0.995} & \dl{\ensuremath{+}0.006}
& 5.53 & \dl{\ensuremath{+}0.04}
& 0.722 & \dl{\ensuremath{+}0.031}
& 3.16 & \dl{\ensuremath{+}0.01}
& 0.742 & \dl{\ensuremath{-}0.027} \\

\midrule

$L_{3\text{-div}}$
& \checkmark & \checkmark & \checkmark
& \checkmark & -- & $L_2$
& 0.989 & \dl{\ensuremath{-}0.006}
& \textbf{5.78} & \dl{\ensuremath{+}0.25}
& 0.726 & \dl{\ensuremath{+}0.004}
& 3.34 & \dl{\ensuremath{+}0.18}
& 0.720 & \dl{\ensuremath{-}0.022} \\

$L_{3\text{-ord}}$
& \checkmark & \checkmark & \checkmark
& -- & \checkmark & $L_2$
& 0.979 & \dl{\ensuremath{-}0.016}
& 5.08 & \dl{\ensuremath{-}0.45}
& \textbf{0.756} & \dl{\ensuremath{+}0.034}
& 3.20 & \dl{\ensuremath{+}0.04}
& 0.737 & \dl{\ensuremath{-}0.005} \\

$L_3$
& \checkmark & \checkmark & \checkmark
& \checkmark & \checkmark & $L_2$
& 0.984 & \dl{\ensuremath{-}0.011}
& 5.59 & \dl{\ensuremath{+}0.06}
& 0.745 & \dl{\ensuremath{+}0.023}
& \textbf{3.35} & \dl{\ensuremath{+}0.19}
& 0.726 & \dl{\ensuremath{-}0.016} \\

\bottomrule
\end{tabular}
}
\end{table}

%% file: tables/05_single_step_round_trip_model_variations_2tab.tex
\begin{table*}[!t]
\centering

\begin{minipage}[t]{0.56\linewidth}
\centering
\caption{Effect of the single-step model on USPTO-190 with $N_B=50$.
Each planner is paired with the template-based LocalRetro in place of the template-free R-SMILES used in Table~\ref{tab:main_results}.
Best and second-best are \textbf{bold} and \underline{underlined}.}
\label{tab:single_step_comparison}

\scriptsize
\setlength{\tabcolsep}{4pt}
\renewcommand{\arraystretch}{1.1}

\begin{adjustbox}{max width=\linewidth}
\begin{tabular}{@{}lcc@{\hspace{6pt}}cccc@{}}
\toprule
\multirow{2}{*}{\textbf{Method}}
& \multicolumn{2}{c}{\textbf{Search Outcome}}
& \multicolumn{4}{c}{\textbf{Route Quality}} \\
\cmidrule(lr){2-3}
\cmidrule(lr){4-7}
& \textbf{Succ. $\uparrow$}
& \textbf{Routes $\uparrow$}
& \textbf{BTR $\uparrow$}
& \textbf{Plaus. $\uparrow$}
& \textbf{RT $\uparrow$}
& \textbf{Price $\downarrow$} \\
\midrule
MORetro*
& 55.79\% & \textbf{20.99} & \underline{43.16\%} & 53.41\% & \underline{44.96\%} & 326.36 \\
RetroAgent
& \textbf{88.95\%} & 4.08 & 38.42\% & \underline{74.29\%} & 32.96\% & \textbf{195.36} \\
\textsc{RetroPlanner}
& \underline{83.16\%} & \underline{5.06} & \textbf{53.16\%} & \textbf{79.45\%} & \textbf{50.38\%} & \underline{207.95} \\
\bottomrule
\end{tabular}
\end{adjustbox}
\end{minipage}
\hfill
\begin{minipage}[t]{0.40\linewidth}
\centering
\caption{\textbf{BTR} and \textbf{RT} on USPTO-190 with RXNGraphormer as the round-trip forward model in place of the ReactionT5v2 used in Table~\ref{tab:main_results}. Parentheses give the percentage-point change.}
\label{tab:round_trip_variation}

\scriptsize
\setlength{\tabcolsep}{4pt}
\renewcommand{\arraystretch}{1.1}

\begin{adjustbox}{max width=\linewidth}
\begin{tabular}{@{}lcc@{\hspace{8pt}}cc@{}}
\toprule
\multirow{2}{*}{\textbf{Method}}
& \multicolumn{2}{c}{\textbf{BTR $\uparrow$}}
& \multicolumn{2}{c}{\textbf{RT $\uparrow$}} \\
\cmidrule(lr){2-3}
\cmidrule(lr){4-5}
& \textbf{RXNG} & $\boldsymbol{\Delta}$
& \textbf{RXNG} & $\boldsymbol{\Delta}$ \\
\midrule
MORetro*    & \underline{50.53\%} & $-1.58$  & \underline{47.17\%} & $-11.77$ \\
RetroAgent  & 46.32\% & $-0.52$  & 38.03\% & $-0.13$ \\
\textsc{RetroPlanner}
& \textbf{64.21\%} & $-2.63$
& \textbf{58.59\%} & $-2.52$ \\
\bottomrule
\end{tabular}
\end{adjustbox}
\end{minipage}

\end{table*}

%% file: contents/05_Limitations_Conclusion.tex
\section{Limitations and Conclusion}

We introduced \textsc{RetroPlanner}, an LLM-based retrosynthesis planner that formulates frontier expansion as a joint decision over multiple molecules rather than committing to a single molecule at each step.
By combining multi-route expansion with chemical evidence-grounded multi-objective reasoning, the planner compares competing routes using reaction plausibility, round-trip consistency, and starting-material price,  while specifying how candidate routes differ from previously explored alternatives. 
Across both in-distribution and out-of-distribution benchmarks, \textsc{RetroPlanner} recovers a broader set of synthesis routes while maintaining a high success rate and improving the quality of the retained alternatives. 
Ablations further show that both the multi-route expansion strategy and the reasoning schema contribute to this behavior, and that the learned planning policy is robust to the choice of single-step and round-trip consistency models.
Our evaluation operationalizes synthesizability through three measurable proxies rather than capturing experimental feasibility in full.
Future work could optimize these route-level objectives directly through reinforcement learning, with credit assignment across multi-step and multi-route decisions, and extend them to laboratory factors such as reaction conditions, yield, and safety.

%% file: contents/99_Appendix.tex
\appendix
\section{Appendix}

\input{contents/88_Related_works}

\subsection{Single-step Models and Chemistry Tools}
\label{appx: tools description}

\paragraph{R-SMILES.}
R-SMILES~\citep{zhong2022root} is a template-free single-step model that translates a product SMILES into reactant-set SMILES using a transformer architecture. Given a target molecule, R-SMILES serializes the product by a depth-first traversal from a chosen root atom, tokenizes the string, and emits a reactant-set string. Since the reactant is serialized from the same root atom, the input and output sequences stay closely aligned, and the decoder edits the string locally instead of regenerating the molecule. Beam search yields $K$ candidate reactant sets per product, each with its sequence likelihood. We use R-SMILES as a single-step model available to the planner.

\paragraph{LocalRetro.}
LocalRetro~\citep{chen2021deep} is a template-based single-step model built on the observation that the atom and bond changes of a reaction are mostly local. LocalRetro extracts local templates that record only the edits at the reaction center, then scores every template at every atom and bond of the product graph. Applying the highest-scoring templates at predicted centers yields precursor sets ranked by scores. We take the top $K$ as the candidate set for the planner.

\paragraph{Reaction Plausibility Filter.}
AiZynthfinder~\citep{genheden2020aizynthfinder} is open-source software for retrosynthetic planning that couples a Monte Carlo tree search with a template-based expansion policy. The expansion policy takes a product fingerprint and outputs a probability over a library of reaction templates. Each template is a SMARTS rule extracted by atom mapping from reactions reported in the USPTO corpus~\citep{Lowe2017}, on which the policy is also trained.  We repurpose the expansion policy of AiZynthfinder as a reaction plausibility filter. 

\paragraph{ReactionT5v2.}
ReactionT5v2~\citep{sagawa2025reactiont5} is a forward reaction prediction model that treats reaction prediction as a sequence-to-sequence translation task. Given a set of reactants as a SMILES string, the model predicts the SMILES of the resulting product. ReactionT5v2 is trained on the Open Reaction Database using a T5 encoder-decoder. We use ReactionT5v2 to compute round-trip consistency~\citep{schwaller2020predicting}.

\paragraph{RXNGraphormer.}
RXNGraphormer~\citep{xu2025unified} couples a graph neural network with a Transformer and is pre-trained on 13 million reactions, then fine-tuned for forward- and retro-synthesis planning as a sequence generation task. We use its forward-synthesis checkpoint as an alternative forward model for round-trip consistency, since its graph-based architecture differs from the sequence-to-sequence models typically used for this axis.

\paragraph{MolPrice.}
MolPrice~\citep{hastedt2025molprice} is a molecular price prediction model used as a proxy for synthetic accessibility. Given a molecular structure, it predicts the price in USD per mmol, trained on 5.5M purchasable compounds extracted from the Molport catalogue. Since the catalogue contains only easily purchasable molecules, a contrastive objective self-generates price labels for hard-to-synthesize structures so that the model assigns them higher prices. We use the predicted price to quantify the price of the building blocks a route terminates in.

\subsection{Implementation Details}
\label{appx:implementation_details}

\paragraph{Training Data Construction.}
\label{appx:training_data_construction}
We draw 3,208 targets from the PaRoutes~\citep{genheden2022paroutes} training split, with gold-route depths ranging from 2 to 14 steps (median 5, mean 4.89).
Terminal molecules are matched against the PaRoutes training stock by InChIKey skeleton.
To avoid overlap with evaluation, we filter molecules appearing in the USPTO-190~\citep{chen2020retro} reference routes and intermediates appearing in the R-SMILES~\citep{zhong2022root} training data.
A final target-level check finds two targets overlapping with USPTO-190 and none with ChEMBL-1000~\citep{liu2023retrosynthetic}, and removing them leaves 3,206 targets.
Each target is searched once with Retro*-0 under a budget of 300 single-step calls.
We Pareto-rank the searched routes over the three synthesis objectives and retain the first two fronts.
The first front alone often concentrates on routes with similar inexpensive starting-material sets, while including the second front broadens route diversity without substantially relaxing route quality.
From these, up to eight routes with distinct starting-material sets are retained per target.
For Pareto ranking, round-trip consistency uses top-1 recovery, a stricter criterion than the top-5 recovery used in evaluation.

\paragraph{Chemical Evidence.}
Bond changes are obtained from RXNMapper~\citep{schwaller2021extraction} atom mappings and represented with CGRtools~\citep{nugmanov2019cgrtools} as formed, broken, or changed bonds at the reaction center, while RDKit~\citep{Landrum2016RDKit2016_09_4} is used to describe the affected atoms and functional groups.
Reaction classes and names are assigned with Rxn-INSIGHT~\citep{Dobbelaere2024-es}, while template matching separately determines whether the corresponding transformation is applicable to the product and whether it reproduces the proposed precursors. 
Molecular descriptors summarize scaffold, ring, size, and stereochemical structure for each molecule and their changes from product to precursors.
When mapping fails, the result is reported as not measured, never as no bond changed, since conflating the two teaches the model to deny perfectly ordinary transformations.

\paragraph{Hyperparameters.}
\input{tables/99_hyperparams}
Training is performed on 4 NVIDIA B200 GPUs with FSDP and bf16 precision, while all inference experiments are conducted on 8 NVIDIA H100 GPUs.
Full training hyperparameters are reported in~\Cref{tab:hyperparams}.

\paragraph{Training Objective.}
We use token-level cross-entropy on packed multi-turn trajectories, supervising the assistant reasoning and action outputs while masking all input and tool-observation tokens.
The loss is normalized over supervised tokens across data-parallel ranks.

\subsection{Evaluation details}
\label{appx:evaluation_details}

\paragraph{Environments.}
When reaction candidates need to be compared, the planner observes a textual rendering of the input search graph.
Each candidate is annotated with four numerical signals, followed by its precursor SMILES.
$q$ is the confidence of the single-step model.
$p$ is reaction plausibility, given by the AiZynthFinder~\citep{genheden2020aizynthfinder} filter trained on USPTO, and a reaction is counted as plausible when $p \ge 0.05$.
$rt$ is round-trip consistency, reported as the rank at which the ReactionT5v2~\citep{sagawa2025reactiont5} forward model recovers the opened molecule from the proposed precursors.
$c_{\mathrm{prec}}(r)$ is the aggregate precursor price displayed for reaction candidate $r$, denoted by $\$$, computed from the MolPrice~\citep{hastedt2025molprice} predictions $c(m)$ of its precursor molecules by log-sum-exp and normalized to $[0,1]$, where lower is cheaper.
Each precursor's individual price $c(m)$ is shown next to its SMILES, and \texttt{*} indicates that the precursor is in stock.

\paragraph{Evaluation Metrics.}
All metrics are computed over the full benchmark $\mathcal{T}$. 
Let $\mathcal{R}_T$ be the set of routes returned for target $T \in \mathcal{T}$ within the budget $N_B$ whose leaves all lie in the eMolecules stock.
We denote solved targets set $\mathcal{S}=\{T:\mathcal{R}_T\neq\emptyset\}$, and evaluate \emph{Success Rate} following:
\begin{equation}
    \emph{Success Rate}=|\mathcal{S}|/|\mathcal{T}|,
\end{equation}

We denote $r \in \rho$ for the reactions of a route $\rho$, and $\operatorname{leaf}(\rho)$ for terminal molecules of route $\rho$. We use the predictors in ~\Cref{appx: tools description}: the single-step reaction plausibility filter $p(r)$, the round-trip $rt(r)$ at which the forward model recovers the product of $r$, and molecule price predictor $c(m)$ that predicts molecule price. 
For a reaction candidate $r$, we denote by $c_{\mathrm{prec}}(r)$ the aggregate predicted price of its precursor molecules.
We first define the route-level quantities as
\begin{equation}
    u_p(\rho) = \frac{1}{|\rho|}\sum_{r\in \rho}\mathbf{1}[p(r) \ge 0.05], \quad
    u_{rt}(\rho) = \frac{1}{|\rho|}\sum_{r\in \rho}\mathbf{1}[rt(r) \le 5],\quad
    \tilde{u}_c(\rho) = \sum_{m \in \operatorname{leaf}(\rho)}c(m),
\end{equation}
Targets for which no route is found are not excluded but scored as $0$ on plausibility and round-trip, and as a price cap $P_{\max}=\$1000$.

Route quality is evaluated on a single representative route $\rho_T^\star$ per solved target. 
While $u_p$ and $u_{rt}$ are already bounded in $[0,1]$, starting-material prices are heavy-tailed and span a much wider numerical range, so we cap the route price at $P_{\max}$ and normalize it on a logarithmic scale to place the cost objective on a comparable scale:
\begin{equation}
\tilde{u}_c^{\mathrm{cap}}(\rho)
=
\min\!\left(\tilde{u}_c(\rho),P_{\max}\right),
\qquad
\bar u_c(\rho)
=
\frac{
\log\!\left(1+\tilde{u}_c^{\mathrm{cap}}(\rho)\right)
}{
\log(1+P_{\max})
}.
\end{equation}
Here, $\bar u_c(\rho)\in[0,1]$, where lower values correspond to cheaper
routes.
We reverse its direction as
$u_c(\rho)=1-\bar u_c(\rho)$ so that all three objectives follow the
same higher-is-better convention.
The representative route is then selected by equal-weight averaging:
\begin{equation}
s(\rho)
=
\frac{u_p(\rho)+u_{rt}(\rho)+u_c(\rho)}{3},
\qquad
\rho_T^\star
=
\arg\max_{\rho\in\mathcal{R}_T}s(\rho).
\end{equation}

\emph{Plausibility} and \emph{Round-Trip} average the representative route's rates over the benchmark, while price is aggregated geometrically under the same cap:
\begin{equation}
\begin{gathered}
\emph{Plausibility}=\frac{1}{|\mathcal{T}|}\sum_{T\in\mathcal{S}}u_p(\rho^{\star}_{T}),
\quad
\emph{Round-Trip}=\frac{1}{|\mathcal{T}|}\sum_{T\in\mathcal{S}}u_{rt}(\rho^{\star}_{T}),
\\[6pt]
\\
\widetilde u_{c,T}
=
\begin{cases}
\min\!\bigl(\tilde{u}_c(\rho_T^\star),P_{\max}\bigr),
& T\in\mathcal S,\\
P_{\max},
& T\notin\mathcal S
\end{cases} \quad
\mathrm{Price}
=
\left(
\prod_{T\in\mathcal S}
\min\!\bigl(\tilde{u}_c(\rho_T^\star),P_{\max}\bigr)
\cdot
P_{\max}^{|\mathcal T|-|\mathcal S|}
\right)^{1/|\mathcal T|}.
\end{gathered}
\end{equation}

\emph{Better-Than-Ref}:
The other metrics compare planners with one another. This one compares them with chemistry that has already been carried out: the benchmark ships with a published route for each target, and we ask how many targets the planner recovers a route that scores better than the published one. We denote $\rho_T^{\mathrm{ref}}$ for the published route of $T$,
\begin{equation}
    \emph{Better-Than-Ref} = \frac{|\{T \in \mathcal{S}: s(\rho_T^\star) > s(\rho_T^{ref})\}|}{|\mathcal{T}|}.
\end{equation}
An unsolved target, or one whose representative route ties with the reference,
counts as a loss.

\emph{1stCutDiff} measures diversity at the initial retrosynthetic decision.
For each recovered route $\rho$ of target $T$, we identify the reaction whose product is $T$ and define
$\mathrm{Prec}_1(\rho)$ as the set of canonicalized precursor molecules produced by this first disconnection.
Given the set of distinct recovered routes $\mathcal{R}_T$, the target-level score is
\begin{equation}
\emph{1stCutDiff}(T)
=
\frac{1}{\binom{|\mathcal{R}_T|}{2}}
\sum_{\{\rho_i,\rho_j\}\subseteq\mathcal{R}_T}
\mathbb{I}\!\left[
\mathrm{Prec}_1(\rho_i)\neq\mathrm{Prec}_1(\rho_j)
\right].
\end{equation}
Thus, two routes are considered different at the first cut only when they decompose the target into different precursor sets, regardless of how their downstream synthesis steps differ.
We compute \emph{1stCutDiff} only for targets with $|\mathcal{R}_T|\geq2$ and report the macro-average over these targets.
We additionally report \emph{1stCuts}, the number of distinct first-cut precursor sets found for each target,
$\emph{1stCuts}(T)=|\{\mathrm{Prec}_1(\rho):\rho\in\mathcal{R}_T\}|$.
While \emph{1stCutDiff} captures pairwise diversity, \emph{1stCuts} measures the absolute breadth of initial synthesis branches reached by the planner.

\subsection{Additional Results}
\label{appx:additional_results}

\paragraph{Baseline Comparison.}
\input{tables/99_retror1_retroagent}

Following the evaluation protocol of RetroAgent~\citep{zhu2026retroagent}, we compare \textsc{RetroPlanner} with conventional search algorithms and recent LLM-based planners on USPTO-190 and ChEMBL-1000, as reported in~\Cref{tab:budget_success_rate}.
For all methods, we use R-SMILES~\citep{zhong2022root} as the single-step retrosynthesis model and measure the search budget $N_B$ by the number of unique single-step model calls. \emph{Success Rate} is defined as the percentage of targets for which at least one complete route is recovered within the given budget. 
As the budget increases, the success rate of \textsc{RetroPlanner} rises to 98.95\% and 96.50\% at $N_B=500$ on USPTO-190 and ChEMBL-1000, respectively.
RetroAgent attains the highest success rates overall, but \textsc{RetroPlanner} remains close while recovering more routes of higher quality.

\paragraph{Effect of Reasoning on Decisions.}
\input{figtabs/99_reasoning_alignment}

To examine whether the reasoning trace informs the action rather than merely preceding it, we measure the attention that action tokens place on the route rows of the input search graph while varying the content of the reasoning field, as shown on the left of \Figref{fig:reasoning_attention}.
Training alone does not change this attention, and filling the field with meaningless text or with the analysis of another instance raises it only slightly.
Attention to route rows is highest only when the field holds the instance's own analysis, indicating that the action draws on the content of its reasoning rather than on its mere presence.
The \texttt{diverging from} field shows the same grounding at a finer level: as shown on the right of \Figref{fig:reasoning_attention}, its tokens attend most strongly to the route they reference rather than to the current branch, tying the stated route relation to the corresponding part of the search graph.

\paragraph{Qualitative analysis of recovered routes.}
\Cref{fig:step_to_pareto} shows the routes \textsc{RetroPlanner} recovers for a single target, placed in the space of the three route-level objectives, together with one Pareto-optimal route (A) and one dominated route (B) drawn in full.
A and B are both five steps long, yet A is plausible and round-trips on every step at \$104, while B does so on 60\% and 40\% of its steps at \$233.
The two share the first disconnection and separate only later, and the objective that separates the leading candidates changes from step to step: plausibility where the routes first diverge, price at the step that closes A, and round-trip consistency at the step that commits A to its deeper subtree.
B fails at two such steps---a disconnection scoring $p=0.01$ and a plausible step ending on a \$170 building block---neither of which is visible from route length or from reaching purchasable building blocks.

\input{figtabs/99_step_to_pareto}

\subsection{Prompts}

\label{appx:reasoning_prompt}
\input{prompts/99_input_prompt1}
\input{prompts/99_input_prompt2}
\input{prompts/99_reasoning_schema}

%% file: contents/88_Related_works.tex
\subsection{Related Work}
\label{main:related_work}

\paragraph{Computer-aided Synthesis Planning.}
Modern computer-aided synthesis planning typically combines a single-step retrosynthesis model with a multi-step search algorithm~\citep{hong2023retrosynthetic, chen2020retro, liu2023retrosynthetic, xie2022retrograph}. 
Single-step models generate candidate precursor sets for a given product molecule and are commonly trained on USPTO-derived datasets such as USPTO-50K and USPTO-FULL~\citep{Lowe2017}. 
Existing approaches can be broadly categorized into template-based methods, which are interpretable but bounded by the template library, template-free methods, which generalize to unseen transformations at the cost of validity and interpretability, and graph-edit-based methods, which balance the two but depend on accurate atom mapping.
Given candidate reactions from a single-step model, multi-step planners determine which molecules and reactions should be expanded. 
Neural-guided planners commonly adopt Monte Carlo Tree Search~\citep{hong2023retrosynthetic}, while methods such as Retro*~\citep{chen2020retro} formulate retrosynthesis as an A*-like best-first search using learned or estimated node costs. 
Other methods, including RetroGraph~\citep{xie2022retrograph} and PDVN~\citep{liu2023retrosynthetic}, further improve search efficiency through alternative graph representations or by jointly learning the single-step policy and value estimates.
Despite these differences, the central objective of conventional planners remains finding at least one route to purchasable building blocks.

\paragraph{Multi-Objective Retrosynthetic Search.}

Retrosynthetic planning can involve multiple criteria, including reaction plausibility, route length, precursor cost, and other measures of synthetic feasibility, which a single solvable route does not necessarily satisfy. 
Multi-objective search methods address this issue by optimizing several route-level objectives and recovering multiple non-dominated routes rather than a single minimum-cost route~\citep{wang2012multi, ganganath2016multiobjective, hastedt2026feasible}. 
Approaches such as MO-MCTS~\citep{wang2012multi} extend conventional retrosynthetic planning toward Pareto-front exploration, building on multi-objective extensions of classical search algorithms such as NAMOA*~\citep{ganganath2016multiobjective}, while MORetro*~\citep{hastedt2026feasible} combines weighted-sum search with adaptive sampling over the weight vector.
However, the objective space is specified in advance, and diversity is obtained by covering it through weight sampling or Pareto-search procedures rather than by reasoning about which parts of the search space remain unexplored.
Moreover, these objectives are typically assumed to decompose additively over reactions, which constrains the notion of route quality that can be optimized during search.
In contrast, we let the planner reason over the current search state to decide which trade-offs to pursue, rather than enumerating them through repeated searches under sampled weights.

\paragraph{LLM Agents for Retrosynthetic Planning.}

Recent work has explored large language models as planners for retrosynthesis.
Earlier systems use LLMs to directly generate synthesis pathways or to guide the construction of retrosynthetic search structures.
More recent approaches, such as Retro-R1~\citep{liu2026retro}, RetrOrchestrator~\citep{changretrorchestrator}, and RetroAgent~\citep{zhu2026retroagent}, formulate retrosynthesis as an agentic workflow in which the LLM interacts with single-step models or chemistry tools and iteratively selects the next molecule or reaction to explore.
These works demonstrate that LLMs can orchestrate tool-based retrosynthetic search through sequential decision-making. 
However, expansion in these systems is confined to a single frontier molecule per turn: the agent inspects the search state, selects one molecule, and applies one or more candidate reactions to it before repeating the process.
Although the search state itself exposes multiple explored and unexplored branches, existing methods do not explicitly learn to use this global context to coordinate complementary exploration across multiple frontier states. 
Our work focuses on this gap by treating the search frontier as a joint decision context and learning to expand multiple routes while simultaneously considering their chemical feasibility.

%% file: tables/99_hyperparams.tex
\begin{table}[t]
\centering
\caption{Training and decoding hyperparameters for \textsc{RetroPlanner}.}
\label{tab:hyperparams}
\small
\setlength{\tabcolsep}{8pt}
\renewcommand{\arraystretch}{1.05}

\begin{tabular}{@{}ll@{}}
\toprule
\textbf{Hyperparameter} & \textbf{Value} \\
\midrule
\multicolumn{2}{@{}l}{\textit{Training}} \\
Base model & GPT-OSS-20B \\
Train / val trajectories & 3,142 / 64 \\
Epochs & 2 \\
Steps & 429 \\
Batch size & 4 (packed, $\sim$393K tokens) \\
Max sequence length & 98,304 \\
Optimizer & AdamW ($\beta_1{=}0.9$, $\beta_2{=}0.95$) \\
Learning rate & 1e-5 \\
LR scheduler & Cosine, min LR 1e-6 \\
Warm-up steps & 20 \\
Weight decay & 0.1 \\
Gradient clipping & 1.0 \\
\midrule
\multicolumn{2}{@{}l}{\textit{Inference}} \\
Temperature & 0.6 \\
Top-$p$ & 1.0 \\
Max new tokens per turn & 3,000 \\
Context length & 131,072 \\
\midrule
\multicolumn{2}{@{}l}{\textit{Hardware}} \\
Training & 4$\times$B200 \\
Inference & 8$\times$H100 \\
\bottomrule
\end{tabular}
\end{table}

%% file: tables/99_retror1_retroagent.tex
\begin{table}[!t]
\centering
\caption{Success rate (\%) under increasing search budgets on USPTO-190 and ChEMBL-1000, with R-SMILES as the single-step model for all methods.
The search budget $N_B$ is the number of unique single-step model calls.
Best and second-best results are in \textbf{bold} and \underline{underlined}, respectively.}
\label{tab:budget_success_rate}

\resizebox{0.92\textwidth}{!}{%
\begin{tabular}{@{}lcccccc@{}}
\toprule

\multirow{2}{*}{\textbf{Method}}
& \multicolumn{6}{c}{\textbf{Success Rate (\%) at Search Budget $N_B$}} \\
\cmidrule(lr){2-7}

& $\mathbf{N_B=50}$
& $\mathbf{N_B=100}$
& $\mathbf{N_B=200}$
& $\mathbf{N_B=300}$
& $\mathbf{N_B=400}$
& $\mathbf{N_B=500}$ \\
\midrule

\multicolumn{7}{c}{\textbf{USPTO-190}} \\
\midrule

\rowcolor[gray]{0.95}
\multicolumn{7}{c}{\textit{Search Algorithms}} \\
\midrule

MCTS
& 44.74
& 52.11
& 58.95
& 67.89
& 71.58
& 72.11 \\

Retro*-0
& 52.63
& 59.47
& 63.68
& 70.00
& 72.11
& 74.74 \\

Retro*
& 57.89
& 64.21
& 71.58
& 79.47
& 80.53
& 84.74 \\

MORetro*
& 67.89
& 81.58
& 88.95
& 91.58
& 94.74
& 96.32 \\

\midrule

\rowcolor[gray]{0.95}
\multicolumn{7}{c}{\textit{LLM-based Search}} \\
\midrule

Retro-R1
& 82.11
& 93.68
& 96.84
& \underline{98.95}
& \underline{98.95}
& \underline{98.95} \\

RetroAgent
& \textbf{96.84}
& \textbf{100.00}
& \textbf{100.00}
& \textbf{100.00}
& \textbf{100.00}
& \textbf{100.00} \\

RetroPlanner
& \underline{92.11}
& \underline{97.37}
& \underline{98.95}
& \underline{98.95}
& \underline{98.95}
& \underline{98.95} \\

\midrule

\multicolumn{7}{c}{\textbf{ChEMBL-1000}} \\
\midrule

\rowcolor[gray]{0.95}
\multicolumn{7}{c}{\textit{Search Algorithms}} \\
\midrule

MCTS
& 59.30
& 64.30
& 70.60
& 78.10
& 82.20
& 82.40 \\

Retro*-0
& 72.70
& 78.00
& 80.00
& 81.50
& 82.20
& 82.50 \\

Retro*
& 74.90
& 79.40
& 80.50
& 81.70
& 83.00
& 83.40 \\

MORetro*
& 77.80
& 84.30
& 89.40
& 91.30
& 92.40
& 93.00 \\

\midrule

\rowcolor[gray]{0.95}
\multicolumn{7}{c}{\textit{LLM-based Search}} \\
\midrule

Retro-R1
& 82.90
& 89.10
& 91.70
& 92.20
& 92.30
& 92.40 \\

RetroAgent
& \textbf{95.90}
& \textbf{98.10}
& \textbf{99.60}
& \textbf{99.70}
& \textbf{99.90}
& \textbf{99.90} \\

RetroPlanner
& \underline{90.00}
& \underline{92.40}
& \underline{94.80}
& \underline{95.90}
& \underline{96.30}
& \underline{96.50} \\

\bottomrule
\end{tabular}%
}
\end{table}

%% file: figtabs/99_reasoning_alignment.tex
\begin{figure}[!t]
    \centering
    \includegraphics[width=0.88
    \textwidth]{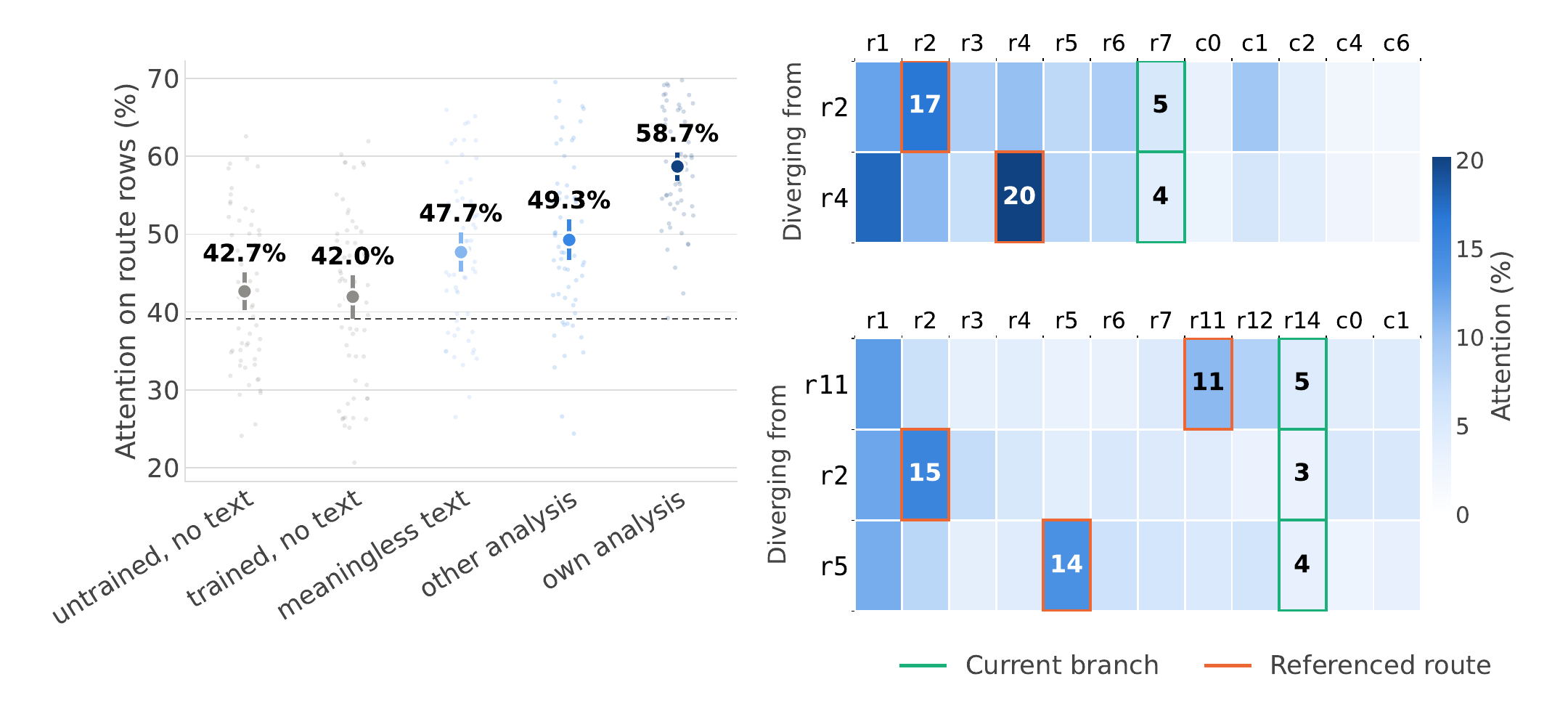}
    \caption{%
    \emph{Left:} Attention from action tokens to the route rows of the input search graph, under different contents of the reasoning field.
    Attention is highest when the field holds the instance's own analysis, compared with no reasoning before or after training, meaningless text, or the analysis of another instance.
    \emph{Right:} Attention from \texttt{diverging from} tokens to the rows of the search graph in two examples.
    Attention concentrates on the referenced route (orange) rather than the current branch (green).%
    }
    \label{fig:reasoning_attention}
\end{figure}

%% file: figtabs/99_step_to_pareto.tex
\begin{figure}[!t]
    \centering
    \includegraphics[width=0.80\textwidth]{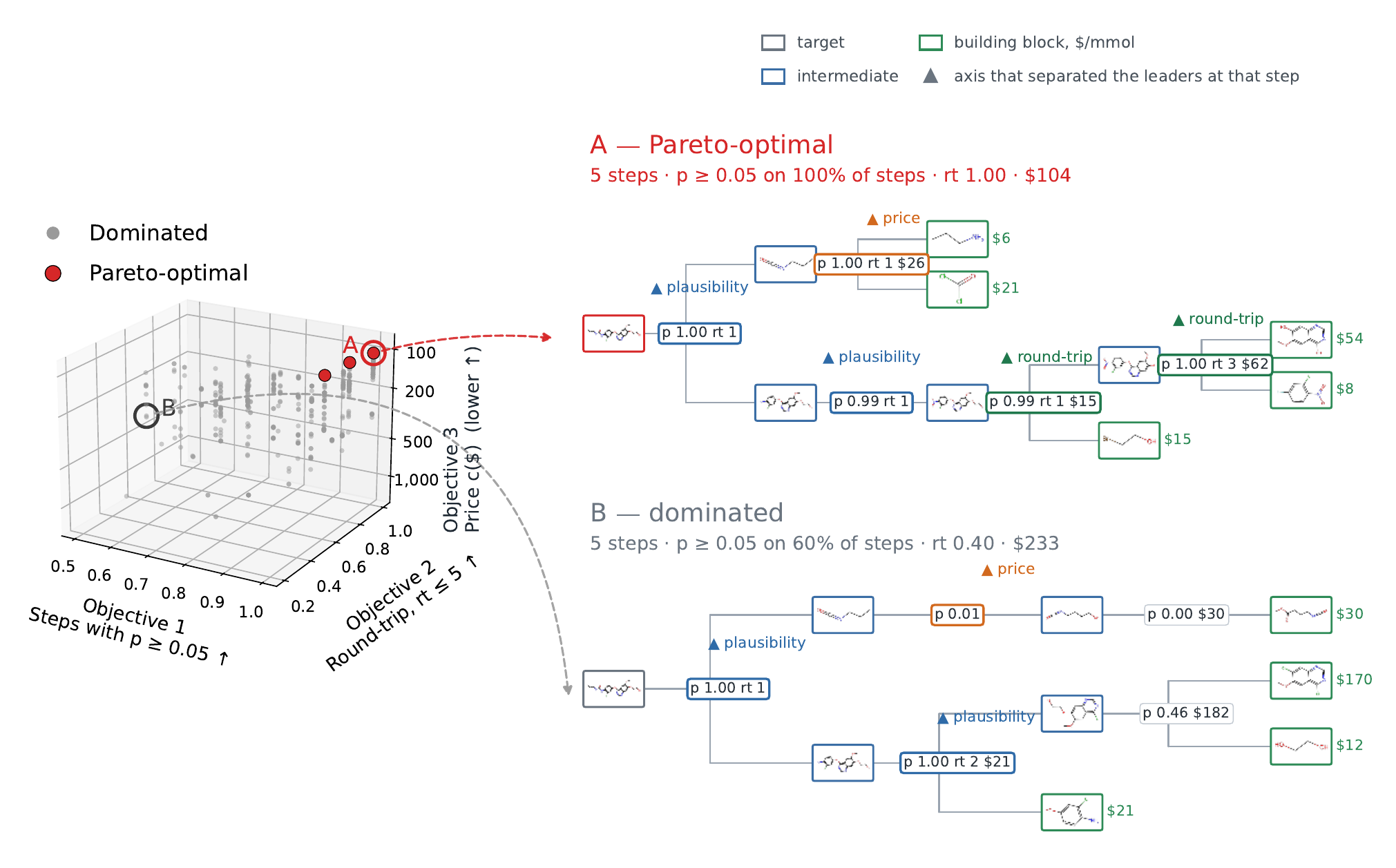}
    \caption{From step-level decisions to route-level objectives for a single target.
    \emph{Left:} Routes recovered by \textsc{RetroPlanner} on three route-level objectives: the fraction of steps with plausibility $p \ge 0.05$, the fraction of steps with round-trip consistency $rt \le 5$, and the total starting-material price $c$, where lower is better.
    Pareto-optimal routes are shown in red and dominated routes in gray.
    \emph{Right:} A Pareto-optimal route (A) and a dominated route (B) for the same target.
    Each reaction is annotated with its $p$, $rt$, and building-block price, and $\blacktriangle$ marks the objective that separated the leading candidates at that step.
    In route A, choices justified by different objectives at each step combine into a route that is not dominated on any objective.}
    \label{fig:step_to_pareto}
\end{figure}

%% file: prompts/99_input_prompt1.tex
\begin{tcolorbox}[colback=pink!5!white, colframe=pink!75!black, title=Input Prompt --- System / Developer / User, breakable]

\fvset{fontsize=\small, breaklines=true, breakanywhere=true,
       breaksymbolleft={}, breaksymbolright={},
       breakanywheresymbolpre={}, breakanywheresymbolpost={}}

\textbf{System}

You are ChatGPT, a large language model trained by OpenAI. Knowledge cutoff: 2024-06. Current date: 2024-06-01. Reasoning: medium. Valid channels: analysis, commentary, final; a channel must be included for every message, and calls to \verb|functions| must go to the commentary channel.

\noindent\rule{\linewidth}{0.2pt}\medskip

\textbf{Developer}

\medskip
\textbf{\# Instructions}

Solve the retrosynthesis search by operating on the board.

\medskip
\textbf{\# Board semantics}

A molecule is an OR: any one reaction under it closes it. A reaction is an AND: every piece under it has to reach purchasable material, or the reaction dies and takes its branch with it. A purchasable piece is closed on arrival. Molecules are shared, so two routes can differ at one reaction and agree everywhere else.

\verb|OPEN| lists every molecule still awaiting a decision. \verb|open| requests the candidate disconnections for a molecule that has no menu. \verb|rank| names the candidates worth taking at a molecule and applies the first \verb|take| of them now (default 1); the rest are the other routes this molecule can give, each one checked to reach purchasable material and carried by number. \verb|done| claims a route: a candidate number for every molecule the route leaves to make, as one object. Handing over is the last message, not a call --- it ends the episode and names the routes you claimed, best first.

Every call carries ONE argument, \verb|actions|, and it is always a list:

\begin{Verbatim}
    {"actions": [{"type": "open", "mid": "u6"}]}
    {"actions": [{"type": "rank", "mid": "u6", "order": [0, 2], "take": 1}]}
    {"actions": [{"type": "done", "choices": {"u6": 0, "n4": 2}}]}
\end{Verbatim}

\verb|take| above 1 expands several subtrees under one molecule in the same call, so more than one of its cuts is live at once. Actions apply in the order you write them, at most one per molecule per call, and a bare action object is not a call. The only actions are \verb|open|, \verb|rank| and \verb|done|: a branch that cannot close is left alone and the board stops offering it. Nothing moves without a call, and the board never repeats a menu you have already been shown.

\medskip
\textbf{\# Board notation}

\verb|<mol e3>SMILES</mol>| is a molecule at first appearance and \verb|<mol e3/>| afterwards, the id two characters assigned once. \verb|c0..c9| are candidate numbers, fixed when the molecule is opened and in the model's own order, so a score is not monotone in the index; outside its own block a candidate is \verb|e3.c1|. \verb|depth D| is how far below the target the piece sits. \verb|rN e3.c1 a of b closed| is the ledger: which candidate was spent, how much of the AND is closed, and the ordering you declared. \verb|CLOSED| is every purchasable leaf banked so far with its price; \verb|ROUTES| lists completed routes together with their route-level scores and terminal leaves.

\medskip
\textbf{\# The signals}

\verb|q| is the single-step model's own confidence in this disconnection; it cannot see what is behind a precursor. \verb|p| is the reaction filter's plausibility with a cut-off of 0.05 --- above it, only this one step is said to go, nothing about where the precursors can go. \verb|rt| is the rank at which the forward model returns the product from these precursors; 1 reads cleanly forwards and \verb|---| means it could not be recovered. \verb|*| marks a fragment already purchasable, saying nothing about cost. \verb|($N)| is what a fragment costs on a 0-to-1 scale, and lower is better --- it is a cost, the opposite direction from \verb|q| and \verb|p|. The line already carries the total: the figure on a cut is computed from its pieces, so quote it rather than one you worked out.

\medskip
\textbf{\# How a route is judged}

\verb|ROUTES| carries three axes for every route claimed, all on the line, and the three are compared without a fixed priority: \verb|p 5/6 pass|, how many steps clear the filter, with \verb|worst at r12| identifying the lowest-plausibility step; \verb|rt 4/6 back|, how many steps the forward model recovers; and \verb|$.218|, what its leaves cost.
\verb|6 steps| is printed too and is not one of the three --- a shorter route can be beaten on all three and still be the one to run, which is a real argument but not a dominance argument. A route beats another when it is at least equal on all three and better on one, over the routes this board has claimed. Two routes that read the same on all three are one answer wearing two labels. Hand over the routes that are not dominated on the three objectives.

\medskip
\textbf{\# Tools}

\begin{Verbatim}
## functions

namespace functions {

// Apply actions to the retrosynthesis board and receive the board that results. Nothing on the board moves without a call.
type board_act = (_: {
  // Actions applied in the order given. At most one per molecule -- so a single call may open several molecules at once, and usually should: the pieces waiting on the board are the frontier, and opening them together brings their disconnections back at the same depth, where the next turn can rank rival cuts against each other instead of committing to one branch and discovering later that it was the wrong one.
  //
  // Every call is wrapped, never a bare action:
  //     {"actions": [ACTION, ...]}
  //
  // Each ACTION is an object:
  //     {
  //       "type":     "open" | "rank" | "done",
  //       "mid"?:     string,                          // open, rank
  //       "order"?:   number[],                        // rank
  //       "take"?:    number,                          // rank, default 1
  //       "choices"?: {molecule id: candidate number}  // done
  //     }
  actions: any[],
}) => any;

} // namespace functions
\end{Verbatim}

\noindent\rule{\linewidth}{0.2pt}\medskip

\textbf{User}

\begin{Verbatim}
OPEN
  m2   depth 0 | no candidates yet
       <mol m2>CN(C)C(=O)c1noc([C@H](CCCC2CCCCC2)CC(=O)NO)n1</mol>
\end{Verbatim}

\end{tcolorbox}

%% file: prompts/99_input_prompt2.tex
\begin{tcolorbox}[colback=pink!5!white, colframe=pink!75!black, title=Rollout Example --- open / rank / done, breakable]

\fvset{fontsize=\small, breaklines=true, breakanywhere=true,
       breaksymbolleft={}, breaksymbolright={},
       breakanywheresymbolpre={}, breakanywheresymbolpost={}}

\textbf{Turn 3 --- assistant (commentary)}

\begin{Verbatim}
{"actions": [{"type": "rank", "mid": "uy", "order": [2, 3, 0], "take": 2},
             {"type": "rank", "mid": "pg", "order": [9, 0, 2], "take": 1}]}
\end{Verbatim}

\textbf{Turn 3 --- tool}

\begin{Verbatim}
[Event] uy ranked [c2, c3, c0] | c2 applied -> r3 | pieces:
          <mol px>CNC</mol>                                                  buyable | $.000
          <mol ba>O=C(C[C@@H](CCCC1CCCCC1)c1nc(C(=O)O)no1)NOC1CCCCO1</mol>   open | depth 2
        r3 1 of 2 closed
[Event] uy ranked [c2, c3, c0] | c3 applied -> r4 | pieces:
          <mol d5>CN(C)C(=O)c1noc([C@H](CCCC2CCCCC2)CC(=O)O)n1</mol>   open | depth 2
          <mol e7>NOC1CCCCO1</mol>                                     buyable | $.076
        r4 1 of 2 closed
[Event] pg ranked [c9, c0, c2] | c9 applied -> r5 | pieces:
          <mol xs>CCI</mol>   buyable | $.014
          <mol d5/>           open | depth 2
        r5 1 of 2 closed

r1  m2.c1  0 of 1 closed        m2 ranked: c1 | c2 | c0
r2  m2.c2  1 of 2 closed        m2 ranked: c1 | c2 | c0
r3  uy.c2  1 of 2 closed        uy ranked: c2 | c3 | c0   (was -)
r4  uy.c3  1 of 2 closed        uy ranked: c2 | c3 | c0   (was -)
r5  pg.c9  1 of 2 closed        pg ranked: c9 | c0 | c2   (was -)

OPEN
  ba   under r3 | depth 2 | no candidates yet
  d5   under r4 | depth 2 | no candidates yet

CLOSED  ea*($.000) | px*($.000) | e7*($.076) | xs*($.014)
\end{Verbatim}

\noindent\rule{\linewidth}{0.2pt}\medskip

\textbf{Turn 4 --- assistant (analysis)}

opening together: ba under r3, d5 under r4 \textbar{} r3 and r4 are 1 of 2 closed, each owing its single open piece; d5 is shared by r4 and r5, so opening both keeps the uy branch's two cuts alive at one depth and lets the next turn rank them side by side.

\textbf{Turn 4 --- assistant (commentary)}

\begin{Verbatim}
{"actions": [{"type": "open", "mid": "d5"}, {"type": "open", "mid": "ba"}]}
\end{Verbatim}

\textbf{Turn 4 --- tool}

\begin{Verbatim}
r1  m2.c1  0 of 1 closed        m2 ranked: c1 | c2 | c0
r2  m2.c2  1 of 2 closed        m2 ranked: c1 | c2 | c0
r3  uy.c2  1 of 2 closed        uy ranked: c2 | c3 | c0
r4  uy.c3  1 of 2 closed        uy ranked: c2 | c3 | c0
r5  pg.c9  1 of 2 closed        pg ranked: c9 | c0 | c2

OPEN
  ba   under r3 | depth 2
    <c0 q.509 p.723 rt2 $1.000>COC(=O)c1noc([C@H](CCCC2CCCCC2)CC(=O)NOC2CCCCO2)n1($1.000)</c0>
    <c1 q.460 p.619 rt5 $.955>CCOC(=O)c1noc([C@H](CCCC2CCCCC2)CC(=O)NOC2CCCCO2)n1($.955)</c1>
    <c2 q.006 p.993 rt- $.634>COC(=O)C[C@@H](CCCC1CCCCC1)c1nc(C(=O)O)no1($.633) + NOC1CCCCO1*($.076)</c2>
    <c3 q.006 p.978 rt2 $.654>NOC1CCCCO1*($.076) + O=C(O)C[C@@H](CCCC1CCCCC1)c1nc(C(=O)O)no1($.653)</c3>
    <c4 q.006 p.619 rt- $.955>CCOC(=O)c1noc(C(CCCC2CCCCC2)CC(=O)NOC2CCCCO2)n1($.955)</c4>
    <c5 q.005 p.723 rt- $1.000>COC(=O)c1noc([C@@H](CCCC2CCCCC2)CC(=O)NOC2CCCCO2)n1($1.000)</c5>
    <c6 q.005 p.000 rt1 $1.000>CC(C)(C)OC(=O)c1noc([C@H](CCCC2CCCCC2)CC(=O)NOC2CCCCO2)n1($1.000)</c6>
    <c7 q.001 p.723 rt- $1.000>COC(=O)c1noc(C(CCCC2CCCCC2)CC(=O)NOC2CCCCO2)n1($1.000)</c7>
    <c8 q.001 p.993 rt- $.634>COC(=O)C[C@H](CCCC1CCCCC1)c1nc(C(=O)O)no1($.633) + NOC1CCCCO1*($.076)</c8>
    <c9 q.001 p1.000 rt2 $.521>C1=COCCC1*($.031) + O=C(C[C@@H](CCCC1CCCCC1)c1nc(C(=O)O)no1)NO($.520)</c9>
  d5   under r4 | depth 2
    <c0 q.504 p.024 rt1 $.546>CN(C)C(=O)c1noc([C@H](CCCC2CCCCC2)CC(=O)OC(C)(C)C)n1*($.546)</c0>
    <c1 q.394 p.999 rt1 $.646>COC(=O)C[C@@H](CCCC1CCCCC1)c1nc(C(=O)N(C)C)no1($.646)</c1>
    <c2 q.047 p.933 rt1 $.576>CCOC(=O)c1noc([C@H](CCCC2CCCCC2)CC(=O)O)n1($.575) + CNC*($.000)</c2>
    <c3 q.046 p.985 rt1 $.589>CCOC(=O)C[C@@H](CCCC1CCCCC1)c1nc(C(=O)N(C)C)no1($.589)</c3>
    <c4 q.005 p.979 rt1 $.667>CNC*($.000) + COC(=O)c1noc([C@H](CCCC2CCCCC2)CC(=O)O)n1($.667)</c4>
    <c5 q.001 p.999 rt- $.646>COC(=O)CC(CCCC1CCCCC1)c1nc(C(=O)N(C)C)no1($.646)</c5>
    <c6 q.001 p.998 rt1 $.653>CNC*($.000) + O=C(O)C[C@@H](CCCC1CCCCC1)c1nc(C(=O)O)no1($.653)</c6>
    <c7 q.001 p1.000 rt1 $.575>CNC*($.000) + O=C(O)C[C@@H](CCCC1CCCCC1)c1nc(C(=O)Cl)no1($.575)</c7>
    <c8 q.001 p.024 rt- $.546>CN(C)C(=O)c1noc(C(CCCC2CCCCC2)CC(=O)OC(C)(C)C)n1*($.546)</c8>
    <c9 q.001 p.024 rt- $.546>CN(C)C(=O)c1noc([C@@H](CCCC2CCCCC2)CC(=O)OC(C)(C)C)n1*($.546)</c9>

CLOSED  ea*($.000) | px*($.000) | e7*($.076) | xs*($.014)
\end{Verbatim}

\noindent\rule{\linewidth}{0.2pt}\medskip

\textbf{Turn 5 --- assistant (analysis)}

\verb|done| on m2$\cdot$c1, uy$\cdot$c3, d5$\cdot$c0: the route ends on the purchasable leaves ea, px, e7, xs. d5 is made at r4, not bought, so it is not a leaf.

\textbf{Turn 5 --- assistant (commentary)}

\begin{Verbatim}
{"actions": [{"type": "done", "choices": {"m2": 1, "uy": 3, "d5": 0}}]}
\end{Verbatim}

\end{tcolorbox}

%% file: prompts/99_reasoning_schema.tex
\begin{tcolorbox}[colback=pink!5!white, colframe=pink!75!black, title=Reasoning Schema --- MOLECULE / DISCONNECTIONS / DECIDING, breakable]

\fvset{fontsize=\small, breaklines=true, breakanywhere=true,
       breaksymbolleft={}, breaksymbolright={},
       breakanywheresymbolpre={}, breakanywheresymbolpost={}}


\textbf{Turn 8 --- assistant (analysis)}

\medskip
\textbf{MOLECULE}

target: oxadiazole bearing a dimethylamide, a cyclohexyl-substituted carbon, and a hydroxamic acid; two ring systems, one aromatic. \\
this piece: the same oxadiazole and cyclohexane, plus a methyl ester, a free hydroxyl, and a tetrahydropyranyl-protected hydroxamate; the ester and the pyran are handles this route installed. \\
not in the target: a methyl ester and a tetrahydropyran ring.

\medskip
\textbf{DISCONNECTIONS OF g3}

c0 \textbar{} bond: forms C-N at the hydroxamate carbonyl carbon and the amine nitrogen; breaks C-O at the hydroxyl oxygen \textbar{} reaction: Carboxylic acid with primary amine to amide \textbar{} what it does: unbalanced, fragments of 24 and 8 heavy atoms; the ring skeleton changes. \\
c1 \textbar{} bond: forms C-N at the hydroxamate carbonyl carbon and the amine nitrogen; breaks C-O at the ester oxygen \textbar{} reaction: Aminolysis of esters \textbar{} what it does: unbalanced, fragments of 25 and 8 heavy atoms; the ring skeleton changes. \\
c2 \textbar{} bond: forms C-N at the hydroxamate carbonyl carbon and the amine nitrogen; breaks C-O at the hydroxyl oxygen \textbar{} reaction: Carboxylic acid with primary amine to amide \textbar{} what it does: unbalanced, fragments of 24 and 8 heavy atoms; the ring skeleton changes. \\
c6 \textbar{} bond: forms C-O at the pyran carbon; changes the order of a C-C bond \textbar{} reaction: oxa-Michael addition \textbar{} what it does: unbalanced, fragments of 25 and 6 heavy atoms; the ring skeleton changes.

\medskip
\textbf{DECIDING g3}

route 1 -- c6 \\
\hspace*{1.2em}diverging from: c1, c2 \textbar{} both are amide couplings of the pyran amine \textbar{} this one closes the pyran by oxa-Michael addition, the amide already in place. \\
\hspace*{1.2em}ordered by: round-trip \textbar{} value: 1 \textbar{} it reads back to the product, which the amide cuts do not.

route 2 -- c1 \\
\hspace*{1.2em}diverging from: r2 \textbar{} same step, Aminolysis of esters, Ester with primary amine to amide \textbar{} it re-forms the hydroxamate from the ethyl ester, a split the oxa-Michael route never makes. \\
\hspace*{1.2em}ordered by: confidence \textbar{} value: 0.320 \textbar{} the two amide cuts read the same on the scored axes, so the single-step model's own reading separates them.

route 3 -- c2 \\
\hspace*{1.2em}diverging from: r4 \textbar{} same step, Carboxylic acid with primary amine to amide \textbar{} it re-forms the hydroxamate from the free acid, the other side of the same amide bond. \\
\hspace*{1.2em}ordered by: structure \textbar{} value: none \textbar{} it beats the head on no number; it is the acid-side coupling, the bond the head leaves untouched.

\medskip\noindent\rule{\linewidth}{0.4pt}\medskip

\textbf{Turn 8 --- assistant (commentary)}

\begin{Verbatim}
{"actions": [{"type": "rank", "mid": "g3", "order": [6, 1, 2], "take": 1}]}
\end{Verbatim}

\noindent\centerline{$\cdots$}

\textbf{Turn 21 --- assistant (analysis)}

\medskip
\textbf{MOLECULE}

target: an oxadiazole bearing a dimethyl amide, a cyclohexyl-substituted side chain, and a hydroxamic acid; two ring systems, one aromatic and one aliphatic, one stereocentre. \\
this piece: g3, the oxadiazole core with the cyclohexyl side chain, an ethyl ester in place of the dimethyl amide, and a tetrahydropyranyl-protected hydroxamate; three ring systems, one stereocentre. \\
not in the target: an ethyl ester and a tetrahydropyranyl protecting group, both handles to be consumed.

\medskip
\textbf{DISCONNECTIONS OF g3}

c0 \textbar{} bond: forms C-N at the hydroxamate carbonyl C; breaks C-O at the hydroxyl O \textbar{} reaction: Carboxylic acid with primary amine to amide \textbar{} what it does: splits it into fragments of 24 and 8  heavy atoms, one piece carrying most of it; the ring skeleton changes. \\
c1 \textbar{} bond: forms C-N at the hydroxamate carbonyl C; breaks C-O at the ester O \textbar{} reaction: Aminolysis of esters, Ester with primary amine to amide \textbar{} what it does: splits it into fragments of 25 and 8 heavy atoms, one piece carrying most of it; the ring skeleton changes. \\
c5 \textbar{} bond: forms C-N; breaks C-O \textbar{} reaction: Aminolysis of esters, Ester with primary amine to amide \textbar{} what it does: splits it into fragments of 26 and 8 heavy atoms, one piece carrying most of it; the ring skeleton changes.

\medskip
\textbf{DISCONNECTIONS OF ba}

c0 \textbar{} bond: breaks C-O at the sp3 C bearing the hydroxyl O \textbar{} reaction: Cleavage of methoxy ethers to alcohols, Ester saponification (alkyl deprotection), Ester saponification (methyl deprotection) \textbar{} what it does: one precursor of 30 heavy atoms; the ring skeleton is kept. \\
c8 \textbar{} bond: forms C-N at the hydroxamate carbonyl carbon and the amine nitrogen; breaks C-O at the ester oxygen \textbar{} reaction: Aminolysis of esters, Ester with primary amine to amide \textbar{} what it does: unbalanced, fragments of 24 and 8 heavy atoms; the ring skeleton changes. \\

\medskip
\textbf{DISCONNECTIONS OF d5}

c0 \textbar{} bond: breaks C-O at the sp3 C bearing the hydroxyl O \textbar{} reaction: Deprotection of carboxylic acid, Ester saponification (alkyl deprotection) \textbar{} what it does: one precursor of 28 heavy atoms; the ring skeleton is kept. \\
c5 \textbar{} bond: breaks C-O \textbar{} reaction: Cleavage of methoxy ethers to alcohols, Ester saponification (alkyl deprotection), Ester saponification (methyl deprotection), Ester to carboxylate \textbar{} what it does: one precursor of 25 heavy atoms; the ring skeleton is kept.

\medskip
\textbf{DECIDING g3}

route 1 -- c5 \\
\hspace*{1.2em}diverging from: r2 \textbar{} same step, Aminolysis of esters, Ester with primary amine to amide \textbar{} this one cuts the hydroxamate side, so it opens a different branch than the amide aminolysis already expanded. \\
\hspace*{1.2em}ordered by: structure \textbar{} value: none \textbar{} the tied leaders were separated by the single-step confidence and a missing round-trip, neither a scored axis, so the cut's own bond change is what it is kept for.

route 2 -- c0 \\
\hspace*{1.2em}diverging from: r4 \textbar{} same step, Carboxylic acid with primary amine to amide \textbar{} this one keeps the ester and cuts the acid side, so it is the acid-side partner to the ester-side cut taken above. \\
\hspace*{1.2em}ordered by: round-trip \textbar{} value: 1 \textbar{} the forward model recovers it cleanly, which the ester-side cut above, at rank two, does not.

\medskip
\textbf{DECIDING ba}

route 1 -- c8 \\
\hspace*{1.2em}diverging from: c2 \textbar{} the same amide cut on the epimeric acid \textbar{} c8 keeps the [C@H] centre the target carries, c2 inverts it. \\
\hspace*{1.2em}ordered by: structure \textbar{} value: none \textbar{} against the plausibility-tied c2, the retained stereocentre decides.

\medskip
\textbf{DECIDING d5}

route 1 -- c5 \\
\hspace*{1.2em}diverging from: none \textbar{} nothing on this board has yet cut this acid, so there is no earlier step to differ from. \\
\hspace*{1.2em}ordered by: plausibility \textbar{} value: .999 \textbar{} the filter clears the step, and it is the only untried cut here that reaches the free acid.

\medskip\noindent\rule{\linewidth}{0.4pt}\medskip

\textbf{Turn 21 --- assistant (commentary)}

\begin{Verbatim}
{"actions": [{"type": "rank", "mid": "g3", "order": [5, 0], "take": 2},
             {"type": "rank", "mid": "ba", "order": [8], "take": 1},
             {"type": "rank", "mid": "d5", "order": [5], "take": 1}]}
\end{Verbatim}

\end{tcolorbox}